\documentclass{article}

\PassOptionsToPackage{numbers, compress}{natbib}

     \usepackage[final]{neurips_2023}

\usepackage{algpseudocode}
\usepackage[utf8]{inputenc} 
\usepackage[T1]{fontenc}    
\usepackage{hyperref}       
\usepackage{url}            
\usepackage{booktabs}       
\usepackage{amsfonts}       
\usepackage{nicefrac}       
\usepackage{microtype}      
\usepackage{xcolor}         

\usepackage{fancyhdr}
\usepackage{wrapfig}
\usepackage{adjustbox}

\usepackage{tikz}
\usetikzlibrary{bayesnet}
\usepackage{amsmath}
\usepackage{algorithm2e}

\usepackage{color}
\usepackage{graphicx}
\usepackage{subcaption}
\usetikzlibrary{backgrounds}
\usetikzlibrary{arrows.meta}
\usepackage{lipsum}
\usepackage{tabularx}
\tikzset{%
    dashedarrow/.style = {%
        draw,  > = {Latex[width = 2.5mm, length = 3.5mm, open, fill = white]}, ->
    }%
}%

\newcommand{\tikzPhilippsDPSSM}{
\begin{tikzpicture}[
 thick,
node distance = 0.5cm and 0.5cm,
minimum size=0.8cm,
invisible_node/.style={draw=none},
]

\node[latent]                                (z0) {${z_{1,1}}$};
\node[latent, right=of z0, xshift=0.7cm]                   (z1) {${z_{1,2}}$};
\node[latent, right=of z1, xshift=0.7cm]                   (z2) {${z_{1,3}}$};
\node[invisible_node, right=of z2, xshift=0.3cm]  (zf) {};

\node[obs, below=of z0]                      (o0) {${w_{1,1}}$};
\node[obs, below=of z1]                      (o1) {${w_{1,2}}$};
\node[obs, below=of z2]                      (o2) {${w_{1,3}}$};
\node[obs, right=of o0,yshift=0.5cm, xshift=-0.5cm, fill={rgb:red,0;green,1;white,3}]                      (a1) {${a_{1,1}}$};
\node[obs, right=of o1,yshift=0.5cm, xshift=-0.5cm, fill={rgb:red,0;green,1;white,3}]                      (a2) {${a_{1,2}}$};

\node[latent, above=of z1]                   (l0)  {${l_1}$};
\node[obs,    above=of l0, yshift=.5cm]                    (c0)  {$\beta_{{1,t}}$};
\node[invisible_node] (inv)    [above = of c0, xshift=.2cm, yshift=-0.7cm]                {$t=1..H$};
\node[latent,    above=of l0,  xshift=-1.5cm, yshift=-0.45cm,]                    (al0)  {$\alpha_{{1}}$};

\edge{z0}{z1}
\edge{z1}{z2}
\edge{z0}{o0}
\edge{a1}{z1}
\edge{a2}{z2}
\edge{z1}{o1}
\edge{z2}{o2}

\edge{l0}{z0}
\edge{l0}{z1}
\edge{l0}{z2}
\edge{l0}{c0}
\edge{al0}{l0}
\edge[dotted]{z2}{zf}

\node[latent, right=of z2,  xshift=2.5cm]                                (z0) {${z_{2,1}}$};
\node[latent, right=of z0, xshift=0.7cm]                   (z1) {${z_{2,2}}$};
\node[latent, right=of z1, xshift=0.7cm]                   (z2) {${z_{2,3}}$};
\node[invisible_node, right=of z2, xshift=0.3cm]  (zf) {};
\node[invisible_node, left=of z0, xshift=-0.3cm]  (zs) {};

\node[obs, below=of z0]                      (o0) {${w_{2,1}}$};
\node[obs, below=of z1]                      (o1) {${w_{2,2}}$};
\node[obs, below=of z2]                      (o2) {${w_{2,3}}$};
\node[obs, right=of o0,yshift=0.5cm, xshift=-0.5cm, fill={rgb:red,0;green,1;white,3}]                      (a1) {${a_{2,1}}$};
\node[obs, right=of o1,yshift=0.5cm, xshift=-0.5cm, fill={rgb:red,0;green,1;white,3}]                      (a2) {${a_{2,2}}$};

\node[latent, above=of z1]                   (l1)  {${l_2}$};
\node[obs,    above=of l1, yshift=.5cm]                    (c1)  {$\beta_{{2,t}}$};
\node[invisible_node] (inv)    [above = of c1, xshift=0.2cm, yshift=-0.7cm]                {$t=1..H$};
\draw[draw=black, rounded corners] (1.41, 4.8) rectangle ++(1.93,-1.8);
\draw[draw=black, rounded corners] (10.51, 4.8) rectangle ++(1.93,-1.8);
\draw[draw=black, rounded corners] (19.61, 4.8) rectangle ++(1.93,-1.8);
\node[latent,    above=of l1,  xshift=-1.5cm, yshift=-0.45cm,]                      (al1)  {$\alpha_{{2}}$};

\edge{z0}{z1}
\edge{z1}{z2}
\edge{z0}{o0}
\edge{a1}{z1}
\edge{a2}{z2}
\edge{z1}{o1}
\edge{z2}{o2}

\edge{l1}{z0}
\edge{l1}{z1}
\edge{l1}{z2}
\edge{l1}{c1}
\edge{al1}{l1}
\edge[dotted]{z2}{zf}
\edge[dotted]{zs}{z0}

\node[latent, right=of z2,  xshift=2.5cm]                                (z0) {${z_{3,1}}$};
\node[latent, right=of z0, xshift=0.7cm]                   (z1) {${z_{3,2}}$};
\node[latent, right=of z1, xshift=0.7cm]                   (z2) {${z_{3,3}}$};
\node[invisible_node, right=of z2, xshift=0.3cm]  (zf) {};
\node[invisible_node, left=of z0, xshift=-0.3cm]  (zs) {};

\node[obs, below=of z0]                      (o0) {${w_{3,1}}$};
\node[obs, below=of z1]                      (o1) {${w_{3,2}}$};
\node[obs, below=of z2]                      (o2) {${w_{3,3}}$};
\node[obs, right=of o0,yshift=0.5cm, xshift=-0.5cm, fill={rgb:red,0;green,1;white,3}]                      (a1) {${a_{3,1}}$};
\node[obs, right=of o1,yshift=0.5cm, xshift=-0.5cm, fill={rgb:red,0;green,1;white,3}]                      (a2) {${a_{3,2}}$};

\node[latent, above=of z1]                   (l2)  {${l_3}$};
\node[obs,    above=of l2, yshift=.5cm]                    (c2)  {$\beta_{{3,t}}$};
\node[latent,    above=of l2,  xshift=-1.5cm, yshift=-0.45cm,]                      (al2)  {$\alpha_{{3}}$};
\node[invisible_node] (inv)    [above = of c2, xshift=0.2cm, yshift=-0.7cm]                {$ t=1..H$};

\edge{z0}{z1}
\edge{z1}{z2}
\edge{z0}{o0}
\edge{a1}{z1}
\edge{a2}{z2}
\edge{z1}{o1}
\edge{z2}{o2}

\edge{l2}{z0}
\edge{l2}{z1}
\edge{l2}{z2}
\edge{l2}{c2}
\edge{al2}{l2}
\edge[dotted]{z2}{zf}
\edge[dotted]{zs}{z0}

\node[invisible_node, right=of l2, xshift=1.5cm]  (lf) {};
\edge{l0}{l1}
\edge{l1}{l2}
\edge[dotted]{l2}{lf}
\end{tikzpicture}
 }

\newcommand{\tikzActAb}{
\begin{tikzpicture}[
thick,
node distance = 0.3cm and 0.3cm,
invisible_node/.style={draw=none,
                       minimum size=0.6cm},
]

\node[obs, minimum size=1cm]                      (r)      {$\alpha_{k,t}$};
\node[obs, minimum size=1cm, below = of r, yshift=2.7cm, xshift=-1.5cm ]        (t)    {$t$};
\node[obs, minimum size=1cm, below = of r, yshift=1cm, xshift=-1.5cm]      (a_t)  {$a_{t,k}$};

\node[latent, minimum size=1cm, right= of r, xshift=0.5cm]         (l)      {$\alpha_k$};

\edge[-{Triangle[open]}] {a_t}{r};
\edge[-{Triangle[open]}] {t}{r};

\edge{l}{r};

\node[invisible_node] (inv)    [above = of a_t, xshift=1.7cm, yshift=1cm]                {$N$};
\draw[draw=black, rounded corners] (-2.2, 1.3) rectangle ++(3.2,-3);

\end{tikzpicture}
}

\newcommand{\tikzTaskinfer}
{ \begin{tikzpicture}[
thick,
node distance = 0.3cm and 0.3cm,
invisible_node/.style={draw=none,
                       minimum size=0.6cm},
]

\node[obs,minimum size=22pt] (r) {$r_{i}$}; 
\node[latent,right=of r, yshift=0cm,xshift=.5cm, minimum size=22pt] (l) { $\ell$};
     
\plate {}{(r)}{$N$};
\edge {l}{r}

     \end{tikzpicture}
}
\newcommand{\cvec}[1]{\boldsymbol{#1}}
\newcommand{\cmat}[1]{\mathbf{#1}}

\usepackage{pifont}

\newcommand{\xmark}{\text{\ding{55}}}

\newtheorem{GU}{Gaussian Update Rule}

\newtheorem{coro}{Corollary}

\newtheorem{minv}{Inversion Of Block Diagonal Matrix}

\title{Multi Time Scale World Models}

\author{%
Vaisakh Shaj$^{1}$\thanks{Corresponding author. Email to <vaisakhs.shaj@gmail.com>.} \quad Saleh Gholam Zadeh$^{1,2}$ \quad Ozan Demir$^3$ \quad Luiz Ricardo Douat$^3$ \\ \textbf{Gerhard Neumann}$^1$ \\
$^1$Karlsruhe Institute Of Technology (KIT), Germany \\ $^2$SAP SE, Germany \\ $^3$Robert Bosch GmbH, Germany\\
}

\begin{document}
\maketitle
\begin{abstract}
Intelligent agents use internal world models to reason and make predictions about different courses of their actions at many scales~\cite{pezzulo2016navigating}. Devising learning paradigms and architectures that allow machines to learn world models that operate at multiple levels of temporal abstractions while dealing with complex uncertainty predictions is a major technical hurdle~\cite{lecun2022path}. In this work, we propose a probabilistic formalism to learn multi-time scale world models which we call the Multi Time Scale State Space (MTS3) model. Our model uses a  computationally efficient inference scheme on multiple time scales for highly accurate long-horizon predictions and uncertainty estimates over several seconds into the future. Our experiments, which focus on action conditional long horizon future predictions, show that MTS3 outperforms recent methods on several system identification benchmarks including complex simulated and real-world dynamical systems. Code is available at this repository: \url{https://github.com/ALRhub/MTS3}.
\end{abstract}
\section{Introduction}

World models attempt to learn a compact and expressive representation of the environment dynamics from observed data. These models can predict possible future world states as a function of an imagined action sequence and are a key ingredient of model-predictive control~\cite{camacho2013model} and model-based reinforcement learning (RL). One important dimension of world models is the level of temporal granularity or the time scale at which the model operates.  Existing literature on world models operates at a single level of temporal abstraction, typically at a fine-grained level such as milliseconds. One drawback of single-time scale world models is that they may not capture longer-term trends and patterns in the data~\cite{lecun2022path}.
 \par
For efficient long-horizon prediction and planning, the model needs to predict at multiple levels of temporal abstractions~\cite {sutton1995td,precup1997multi}.  Intuitively, low-level temporal abstractions should contain precise details about the input so as to predict accurately in the short term, while high-level, abstract representations should simplify accurate long-term predictions. Both abstractions must also interrelate with each other at least in the sense that the higher-level predictions/plans can be turned into low-level moment-by-moment predictions. 
For example, in robotic manipulation, the robot must be able to perform precise and coordinated movements to grasp and manipulate the object at a fast time scale while at a slower time scale, the robot must also be able to recognize and utilize higher-level patterns and structures in the task, such as the shape, size and location of objects, and the overall goal of the manipulation task. \par
Furthermore, temporal abstractions can capture relevant task structures across dynamical systems under non-stationary which can be used to identify the similarities and differences between tasks, allowing the robot to transfer knowledge learned from one task to another \cite{shanahan2022abstraction, lecun2022path}. 
\par
In this work, we attempt to come up with a principled probabilistic formalism for learning such multi-time scale world models as a hierarchical sequential latent variable model. We show that such models can better capture the complex, non-linear dynamics of a system more efficiently and robustly than models that learn on a single timescale. This is exemplified in several challenging simulated and real-world prediction tasks such as the D4RL dataset, a simulated mobile robot and real manipulators including data from heavy machinery excavators. 

\section{Preliminaries}
State space models (SSMs) are Bayesian probabilistic graphical models \citep{koller2009probabilistic,jordan2004graphical} that are popular for learning patterns and predicting behaviour in sequential data and dynamical systems.
Formally, we define a state space model as a tuple $( \mathcal{Z}, \mathcal{A}, \mathcal{O}, f, h, \Delta t)$, where $\mathcal{Z}$ is the state space, $\mathcal{A}$ the action space and $\mathcal{O}$ the observation space of the SSM. The parameter $\Delta t$ denotes the discretization time-step and $f$ and $h$ the dynamics and observation models respectively.  
We will consider the Gaussian state space model that is represented using the following equations
\begin{equation*}
  \begin{aligned}[c]
  \cvec{z}_t &= f(\cvec{z}_{t-1},\cvec{a}_{t-1}) + \cvec{\epsilon}_t, \quad \cvec{\epsilon}_t \sim \mathcal{N}(\cvec{0}, \cmat{\Sigma_z}),
  \end{aligned} \textrm{and}
  \quad
  \begin{aligned}[c]
   \cvec{o}_t &= h(\cvec{z}_t) + \cvec{v}_t, \quad \cvec{v}_t \sim \mathcal{N}(\cvec{0}, \cmat{\Sigma_o}).
  \end{aligned}
\end{equation*}
Here $\cvec{z}_t \in \mathcal{Z}$, $\cvec{a}_t \in \mathcal{A}$ and $\cvec{o}_t \in \mathcal{O}$ are the latent states, actions and observations at time t. The vectors $\cvec{\epsilon}_t$ and $\cvec{v}_t$ denote zero-mean Gaussian noise.
When $f$ and $h$ are linear/locally linear, inference can be performed efficiently via exact inference. There have been several works recently~\cite{haarnoja2016backprop,becker2019recurrent, shaj2022hidden} where these closed-form solutions are coded as layers of a neural network in deep-state space model literature, i.e, the architecture of the network is informed by the structure of the probabilistic state estimator. Following this line of work, we propose a multi-time scale linear Gaussian state space model, whose inference can be performed via closed-form solutions.

\paragraph{(Locally-)Linear Gaussian SSMs.}
To perform inference in SSMs, we follow \cite{becker2019recurrent}. They use a (locally-)linear dynamics model. Moreover, they replace the observations $\cvec o$ with a latent observation $\cvec w$. This latent observation is obtained by an encoder $\cvec w_{ o_t} = \textrm{enc}_w(\cvec o_{t})$ along with the uncertainty of this observation, i.e., $\cvec \sigma_{o_t} = \textrm{enc}_{\sigma}(\cvec o_{t}).$ Due to the non-linear observation encoder, a simplified linear observation model can now be used. Hence, the dynamics and observation models can be described as
$$p(\cvec z_{t+1}|\cvec z_t, \cvec a_t) =  \mathcal{N}(\cvec A_t \cvec z_t + \cvec c_t + \cvec B_t \cvec a_t,\textrm{diag}(\cvec \sigma_z)), \textrm{ and } p(\cvec w_{o_t}|\cvec z_t) =  \mathcal{N}(\cvec H \cvec z_t,\textrm{diag}(\cvec \sigma_{o_t})),$$
where a simple observation matrix of $\cvec H = [\cvec I,  \cvec 0]$ is used. The underlying assumption behind this observation model is that the latent state $\cvec z_t = [\cvec p_t^T, \cvec d_t^T]^T$ has twice the dimensionality of the latent observation $\cvec w_t$ and only the first half of the latent state, i.e., $\cvec p_t$, can be observed. The second half of the latent state, i.e., $\cvec d_t$, serves as derivative or velocity units that can be used by the model to estimate the change of the observable part of the latent state. 

\paragraph{Factorized Inference in Linear Gaussian SSMs.}
Inference in the introduced linear Gaussian SSM is straightforward and can be performed using Kalman prediction and observation updates. However, these updates involve high dimensional matrix inversions that are expensive to evaluate and hard to backpropagate for end-to-end learning. Hence, \cite{becker2019recurrent} introduce a factorization of the belief $p(\cvec z_t|\cvec o_{1:t}, \cvec a_{1:t-1}) = \mathcal{N}(\cvec \mu_t, \cvec \Sigma_t)$ such that only the diagonal and one off-diagonal vector of the covariance need to be computed, i.e. 

$$\cvec \Sigma_t = \left[\begin{array}{cc}\cvec \Sigma_t^u & \cvec \Sigma^{s}_t \\ \cvec \Sigma^{s}_t & \cvec \Sigma^{l}_t  \end{array} \right], \textrm{ with } \cvec \Sigma_u = \textrm{diag}(\cvec \sigma_t^s), \; \cvec \Sigma_l = \textrm{diag}(\cvec \sigma_t^l) \textrm{ and } \cvec \Sigma_{s} = \textrm{diag}(\cvec \sigma_{t}^s).$$
Using this factorization assumption, closed-form Gaussian inference can be performed using only scalar divisions which are fast and easy to back-propagate. These factorization assumptions form the basis for the inference update in our MTS3 model. 

\paragraph{Bayesian Aggregation.} To aggregate information from several observations into a consistent representation, \cite{volpp2020bayesian} introduce Bayesian aggregation in the context of Meta-Learning. They again use an encoder to obtain a latent observation vector $\cvec r_{o_t}$ and its uncertainty vector $\cvec \sigma_{o_t}$. Given the observation model $p(\cvec r_{o_t}|\cvec z) = \mathcal{N}(\cvec H \cvec z, \textrm{diag}(\cvec \sigma_{o_t}))$ with $\cvec H = \cvec I$ and a prior $p(\cvec z) = \mathcal{N}(\cvec \mu_0, \textrm{diag}(\cvec \sigma_0))$, the posterior $p(\cvec z| \cvec r_{o_1:o_t})$ can again be effectively computed by Gaussian inference that involve only scalar inversions. Note that computing this posterior is a simplified case of the Kalman update rule used in Gaussian SSMs \cite{becker2019recurrent}, with no memory units, $\cvec H 
 = \cvec I$ and no dynamics. To increase efficiency, the update rule can be formulated in a batch manner for parallel processing instead of an incremental update \cite{volpp2020bayesian}. 


\section{Multi Time Scale State Space Models}
\begin{wrapfigure}[12]{r}{.38\linewidth}
 \scalebox{0.75}{
\begin{subfigure}[b]{.38\textwidth}
    \centering
    \includegraphics[scale=0.33]{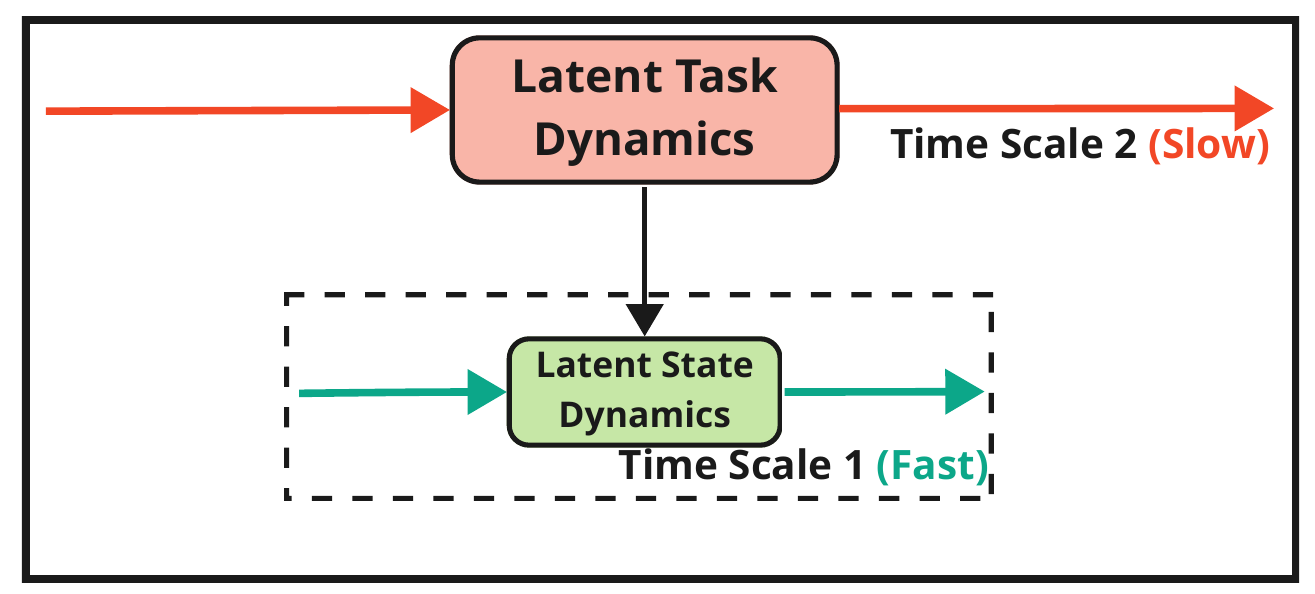}
     \end{subfigure}}
\caption{MTS3 captures slow-moving long-term trends as the latent task dynamics and the fast-moving short-term trends as the latent state dynamics.} 
 \label{fig:pgm}
\end{wrapfigure}
Our goal is to learn a principled sequential latent variable model that can model the dynamics of partially observable robotic systems under multiple levels of temporal abstractions. To do so, we introduce a new formalism, called Multi Time Scale State Space (MTS3) Model, with the following desiderata:
i) It is capable of modelling dynamics at multiple time scales. ii) It allows for a single global model to be learned that can be shared across changing configurations of the environments. iii) It can give accurate long-term predictions and uncertainty estimates. iv) It is probabilistically principled yet scalable during learning and inference.

\subsection{General Definition}
An MTS3 model with 2 timescales is defined by two SSMs on a fast and a slow time scale respectively. Both SSMs are coupled via the latent state of the slow time scale SSM, which parametrizes/``reconfigures'' the system dynamics of the fast time scale SSM. While the fast time scale SSM runs at the original time step $\Delta t$ of the dynamical system, the slow time scale SSM is only updated every $H$ step, i.e., the slow time scale time step is given by $H \Delta t$. We will derive closed-form Gaussian inference for obtaining the beliefs for both time scales, resulting in variations of the Kalman update rule which are also fully differentiable and used to back-propagate the error signal \cite{becker2019recurrent,haarnoja2016backprop}. The definition with a 2-level MTS3 along with the inference and learning schemes that we propose is directly extendable to an arbitrary number of temporal abstractions by introducing additional feudal~\cite{dayan1992feudal} hierarchies with longer discretization steps and is further detailed in Section \ref{subsec: Nlevel}.

\subsubsection{Fast time-scale SSM}
\label{sec: fts}
  The fast time-scale (fts) SSM is given by $\mathcal{S}_{\textrm{fast}} = ( \mathcal{Z}, \mathcal{A}, \mathcal{O}, f_{\cvec l}^\textrm{fts}, h^\textrm{fts}, \Delta t, \mathcal{L}).$ Here, $\cvec l \in \mathcal{L}$ is a task descriptor that parametrizes the dynamics model of the SSM and is held constant for H steps. We will denote the task descriptor for the $k$th time window of $H$ steps as $\cvec l_k$. The probabilistic dynamics and observation model of the fast time scale for the $t$th time step in the $k$th window can then be described as  
 \begin{align}
  p(\cvec{z}_{k,t}|\cvec{z}_{k,t-1},\cvec{a}_{k,t-1}, \cvec l_k) & = \mathcal{N}(f^\textrm{fts}_{\cvec l}(\cvec{z}_{k,t-1},\cvec{a}_{k,t-1}, \cvec l_k), \cmat{Q}),
   \textrm{ and } \nonumber\\
   p(\cvec{o}_{k,t}|\cvec{z}_{k,t}) & = \mathcal{N}(h^\textrm{fts}(\cvec{z}_{k,t}), \cmat{R}).
 \end{align}
 \paragraph{Task-conditioned marginal transition model.}
Moreover, we have to consider the uncertainty in the task descriptor (which will, in the end, be estimated by the slow time scale model), i.e., instead of considering a single task descriptor $\cvec l_k$, we have to consider a distribution over task-descriptors $p(\cvec l_k)$ for inference in the fts-SSM. This distribution will be provided by the slow-time scale SSM for every time window $k$. We can further define the marginal task-conditioned transition model for time window $k$ which is given by 
\begin{align}p_{\cvec l_k}(\cvec{z}_{k,t}|\cvec{z}_{k,t-1},\cvec{a}_{k,t-1}) & = \int p(\cvec{z}_{k,t}|\cvec{z}_{k,t-1},\cvec{a}_{k,t-1}, \cvec l_k) p(\cvec l_k) d \cvec l_k
\end{align}
\paragraph{Latent observations.}
Following \cite{becker2019recurrent}, we replace the observations by latent observations and their uncertainty, i.e., we use latent observation encoders to obtain $\cvec w_{k,t} = \textrm{enc}_w(\cvec o_{k,t})$ and an uncertainty encoder $\cvec \sigma_{k,t} = \textrm{enc}_\sigma(\cvec o_{k,t})$. The observation model is hence given by $p(\cvec w_{k,t}|\cvec z_{k,t}) = \mathcal{N}(h^\textrm{fts}(\cvec z_{k,t}), \textrm{diag}(\cvec \sigma_{k,t}))$.

\subsubsection{Slow time-scale SSM}
\label{sts}
The slow time-scale (sts) SSM only updates every H time step and uses the task parameter $\cvec l$ as latent state representation. Formally, the SSM is defined as $\mathcal{S}_{\textrm{slow}} = ( \mathcal{L}, \mathcal{E}, \mathcal{T}, f^\textrm{sts}, h^\textrm{sts}, H \Delta t)$. It uses an abstract observation  $\cvec \beta \in \mathcal{B}$ and abstract action $\cvec \alpha \in \mathcal{A}$ that summarize the observations and actions respectively throughout the current time window. The general dynamics model is hence given by \begin{align}p(\cvec l_{k}|\cvec l_{k-1}, \cvec \alpha_k) = \mathcal{N}(f^{\textrm{sts}}(\cvec l_{k-1}, \cvec \alpha_k), \cvec S).\end{align}

While there exist many ways to implement the abstraction of observations and actions of the time windows, we choose to use a consistent formulation by fusing the information from all $H$ time steps of time window $k$ using Gaussian conditioning. 
 \begin{figure*}[h]
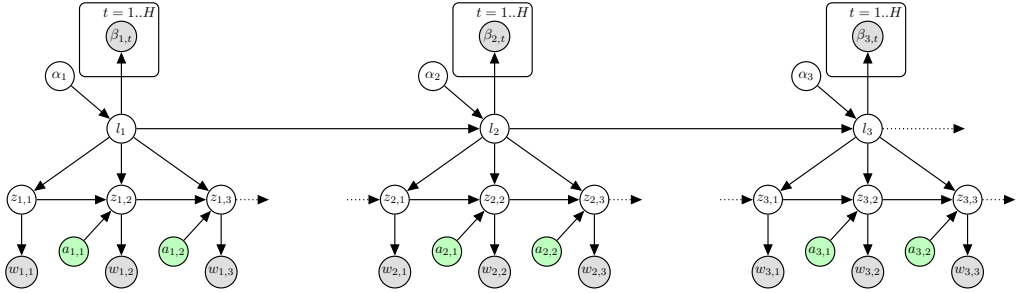

\begin{center}
 \resizebox{\linewidth}{!}{\tikzPhilippsDPSSM}
\end{center}
\caption{The graphical model corresponding to an MTS3 with 2 timescales. The latent task variable $\cvec l_k$ captures the slow-changing dynamics using abstract observation inferred from $\{\cvec \beta_{k,t}\}_{t=1}^H$and abstract action $\cvec \alpha_k$ as described in section \ref{sts}. The inference in the fast time scale uses primitive observations $\cvec w_{k,t}$, primitive actions $\cvec a_{k,t}$ and the latent task descriptor $l_k$ which parameterizes the fast-changing dynamics of $\cvec z_{k,t}$ for a time window k as discussed in the section \ref{sec: fts-inf}.} 
 \label{fig:mts}
\end{figure*}
\paragraph{Observation abstraction.}
In terms of the abstract observation model, we choose to model $H$ observations $\cvec \beta_{k,t}$, $t \in [1,H]$ for a single slow-scale time step $k$. All these observations can then be straightforwardly integrated into the belief state representation using incremental observation updates. The abstract observation and its uncertainty for time step $t$ is again obtained by an encoder architecture, i.e, 
$$\cvec \beta_{k,t} = \textrm{enc}_{\beta}(\cvec o_{k,t}, t), \quad \cvec \nu_{k,t} = \textrm{enc}_{\nu}(\cvec o_{k,t}, t),$$
and $p(\cvec \beta_{k,t}|\cvec l_k) = \mathcal{N}(h^{\textrm{sts}}(\cvec l_k), \textrm{diag}(\cvec \nu_{k,t})).$
Hence, the abstract observation $\cvec \beta_{k,t}$ contains the actual observation $\cvec o_{k,t}$ at time step $t$ as well as a temporal encoding for the time-step. While multiple Bayesian observation updates are permutation invariant, the temporal encoding preserves the relative time information between the observations, similar to current transformer architectures.

\paragraph{Action abstraction.} The abstract action $\cvec \alpha_k$ causes the transitions to the latent task $\cvec l_{k}$ from $\cvec l_{k-1}$. It should contain the relevant information of all primitive actions $\cvec a_{k,t}$, $t \in [1,H]$ executed in the time window $k$. To do so, we again use Bayesian conditioning and latent action encoding. Each control action $\cvec a_{k,t}$ and the encoding of time-step $t$ is encoded into its latent representation and its uncertainty estimate, i.e.,
$$\cvec \alpha_{k,t} = \textrm{enc}_{\alpha}(\cvec a_{k,t}, t), \quad \cvec \rho_{k,t} = \textrm{enc}_{\rho}(\cvec a_{k,t}, t).$$
The single latent actions $\cvec \alpha_{k,t}$ can be aggregated into a consistent representation $\cvec \alpha_{k}$ using Bayesian aggregation \cite{volpp2020bayesian}. To do so, we use the likelihood $p(\cvec \alpha_{k,t}|\cvec \alpha_{k}) = \mathcal{N}(\cvec \alpha_{k}, \textrm{diag}( \cvec \rho_{k,t}))$ and obtain the posterior $p(\cvec \alpha_{k}|\cvec \alpha_{k,1:H}) = \mathcal{N}(\cvec \mu_{ \alpha_{k}}, \cvec \Sigma_{\alpha_k})$, which is obtained by following the standard Bayesian aggregation equations, see Appendix A. Note that our abstract action representation also contains an uncertainty estimate which can be used to express different effects of the actions on the uncertainty of the prediction. Due to the Gaussian representations, we can compute the marginal transition model 
\begin{align}p_{\cvec \alpha_k}(\cvec l_{k}|\cvec l_{k-1}, \cvec \alpha_{k,1:H}) = \int p_{\cvec \alpha_k}(\cvec l_{k}|\cvec l_{k-1}, \cvec \alpha_k) p(\cvec \alpha_k |\cvec \alpha_{k,1:H}) d\cvec \alpha_k.\end{align}
This transition model is used for inference and its parameters are learned. 

\subsubsection{Connecting both SSMs via inference}
In the upcoming sections, we will devise Bayesian update rules to obtain the prior $p(\cvec l_k| \cvec \beta_{1:k-1},  \cvec \alpha_{1:k})$ and posterior $p(\cvec l_k| \cvec \beta_{1:k},  \cvec \alpha_{1:k})$ belief state for the sts-SSM as well as the belief states for the fts-SSM. The prior belief $p(\cvec l_k| \cvec \beta_{1:k-1},  \cvec \alpha_{1:k})$ contains all information up to time window $k-1$ and serves as a distribution over the task-descriptor of the fts-SSM, which connects both SSMs. This connection allows us to learn both SSMs jointly in an end-to-end manner.  

The probabilistic graphical model of our MTS3 model is depicted in Figure \ref{fig:mts}. In the next section, we will present the detailed realization of each SSM to perform closed-form Gaussian inference and end-to-end learning on both time scales.

\subsection{Inference in the Fast Time-Scale SSM}
\label{sec: fts-inf}
The fts-SSM performs inference for a given time window $k$ of horizon length $H$. To keep the notation uncluttered, we will also omit the time-window index $k$ whenever the context is clear. 
We use a linear Gaussian task conditional transition model, i.e, \begin{align}p(\cvec{z}_{t}|\cvec{z}_{t-1},\cvec{a}_{t-1},\cvec{l}_k) = \mathcal{N}\left(\cmat{A}\cvec{z}_{t-1} + \cmat{B}\cvec{a}_{t-1} + \cmat{C}\cvec{l}_k,  \cmat Q\right),\end{align} where $\cvec A$, $\cvec B$, $\cvec C$ and $\cvec Q$ are state-independent but learnable parameters.
In our formulation, the task descriptor can only linearly modify the dynamics which was sufficient to obtain state-of-the-art performance in our experiments, but more complex parametrizations, such as locally linear models, would also be feasible. Following \cite{becker2019recurrent}, we split the latent state $\cvec z_t = [\cvec p_t, \cvec m_t]^T$ into its observable part $\cvec p_t$ and a part $\cvec m_t$ that needs to be observed over time. 
We also use a linear observation model $p(\cvec w_{t}|\cvec z_{t}) = \mathcal{N}(\cvec H \cvec z_{t}, \textrm{diag}(\cvec \sigma_{t}))$ with $\cvec H = [\cvec I, \cvec 0]$. 

We will assume that the distribution over the task descriptor is also given by a Gaussian distribution, i.e., $p(\cvec l_k) = \mathcal{N}(\cvec \mu_{\cvec l_k}, \cvec \Sigma_{\cvec l_k})$, which will be provided by the slow-time scale (sts) SSM, see Section \ref{sts-inf}. Given these modelling assumptions, the  task variable can now be integrated out in closed form, resulting in the following task-conditioned marginal transition model
\begin{align}p_{\cvec l_k}(\cvec{z}_{t}|\cvec{z}_{t-1},\cvec{a}_{t-1}) 
 = \mathcal{N}\left(\cmat{A}\cvec{z}_{t-1} + \cmat{B}\cvec{a}_{t-1} + \cmat{C}\cvec{\mu}_{\cvec l_k},  \cmat{Q} + \cvec C \cvec \Sigma_{\cvec l_k} \cvec C^T \right), \end{align}
which will be used instead of the standard dynamics equations.
 We follow the same factorization assumptions as in \cite{becker2019recurrent} and only estimate the diagonal elements of the block matrices of the covariance matrix of the belief, see Appendix B. The update equations for the Kalman prediction and observation updates are therefore equivalent to the RKN~\cite{becker2019recurrent}.

\subsection{Inference in the Slow-Time Scale SSM}
\label{sts-inf}
\paragraph{Prediction Update.}
We follow the same Gaussian inference scheme as for the fts-SSM, i.e., we again employ a linear dynamics model 
$p(\cvec l_{k}|\cvec l_{k-1}, \cvec \alpha_k) = \mathcal{N}(\cvec X \cvec l_{k-1} + \cvec Y \cvec \alpha_{k}, \cvec S),$ where $\cvec X$, $\cvec Y$ and $\cvec S$ are learnable parameters. The marginalized transition model for the abstract actions is then given by 

\begin{align}p_{\cvec \alpha_k}(\cvec{l}_{k}|\cvec{l}_{k-1}) & = \int p(\cvec{l}_{k}|\cvec{l}_{k-1},\cvec{\alpha}_{k}) p(\cvec \alpha_k) d \cvec \alpha_k 
 = \mathcal{N}\left(\cmat{X}\cvec{l}_{k-1} + \cmat{Y}\cvec{\mu}_{\alpha_k}, \cmat{S} + \cvec Y \cvec \Sigma_{\cvec \alpha_k} \cvec Y^T \right). \end{align}

We can directly use this transition model to obtain the Kalman prediction update which computes the prior belief $p_{\cvec \alpha_{1:k}}(\cvec l_k | \cvec \beta_{1:k-1}) = \mathcal{N}(\cvec \mu_{l_k}^-, \cvec \Sigma_{l_k}^-)$ from the posterior belief  $p_{\cvec \alpha_{1:k-1}}(\cvec l_{k-1} | \cvec \beta_{1:k-1}) = \mathcal{N}(\cvec \mu_{l_{k-1}}^+, \cvec \Sigma_{l_{k-1}}^+)$ of the previous time window, see Appendix A. 

\paragraph{Observation Update.}
Similarly, we will use a linear observation model for the abstract observations 
$p(\cvec \beta_{k,t}| \cvec l_k) = \mathcal{N}(\cmat H \cvec l_{k}, \textrm{diag}(\cvec \nu_{k,t}))$ with $\cmat H = [\cmat I, \cmat 0]$.
As can be seen from the definition of the observation matrix $\cmat H$, the latent space is also decomposed into its observable and unobservable part, i.e., $\cvec l_k = [\cvec u_{k}, \cvec v_{k}]$. In difference to the standard factorized Kalman observation update given in Appendix A, we have to infer with a set of observations $\vec \beta_{k,t}$ with $t = 1 \dots H$ for a single time window $k$. While in principle, the Kalman observation update can be applied incrementally $H$ times to obtain the posterior $p_{\cvec \alpha_{1:k}}(\cvec l_{k} | \cvec \beta_{1:k}) = \mathcal{N}(\cvec \mu_{l_{k}}^+, \cvec \Sigma_{l_{k}}^+)$, such an update would be very slow and also cause numerical inaccuracies. Hence, we devise a new permutation invariant version of the update rule that allows parallel processing with set encoders~\cite{zaheer2017deep}. We found that this update rule is easier to formalize using precision matrices. Hence, we first transform the prior covariance vectors $\cvec \sigma_{l_{k}}^{u-}$, $\cvec \sigma_{l_{k}}^{l-}$ and $\cvec \sigma_{l_{k}}^{s-}$ to its corresponding precision representation $\cvec \lambda_{l_{k}}^{u-}$, $\cvec \lambda_{l_{k}}^{l-}$ and $\cvec \lambda_{l_{k}}^{s-}$ which can be performed using block-wise matrix inversions of $\cvec \Sigma_{l_{k}}^-$. Due to the factorization of the covariance matrix, this operation can be performed solely by scalar inversions. As the update equations are rather lengthy, they are  given in Appendix A, B. Subsequently, we compute the posterior precision, where only $\cvec \lambda_{l_{k}}^{u}$ is changed by 
\begin{align}\cvec \lambda_{l_{k}}^{u+} = \cvec \lambda_{l_{k}}^{u-} + \sum_{t=1}^H \cvec 1 \oslash \cvec \nu_{k,t}\end{align}
 while $\cvec \lambda_{l_{k}}^{l+} = \cvec \lambda_{l_{k}}^{l-}$ and $\cvec \lambda_{l_{k}}^{s+} = \cvec \lambda_{l_{k}}^{s-}$ remain constant. The operator $\oslash$ denotes the element-wise division. From the posterior precision, we can again obtain the posterior covariance vectors $\cvec \sigma_{l_{k}}^{u+}$, $\cvec \sigma_{l_{k}}^{l+}$ and $\cvec \sigma_{l_{k}}^{s+}$ using only scalar inversions, see Appendix A, B. The posterior mean $\cvec{\mu}_{l,k}^{+}$ can now be obtained from the prior mean $\cvec{\mu}_{l,k}^{-}$ as
 \begin{equation}
  \begin{aligned}[c]
   \cvec{\mu}_{l,k}^{+}=\cvec{\mu}_{l,k}^{-} + \left[ \begin{array}{l}
\cvec{\sigma}_{l_k}^{u+} \\
\cvec{\sigma}_{l_k}^{s+} \\
\end{array}\right]  \odot \left[ \begin{array}{l}
\sum_{t=1}^{H}  \left(\cvec{\beta}_{k,t}-\cvec{\mu}^{\mathrm{u},-}_{l_k}\right) \oslash \cvec \nu_{k,t} \\
\sum_{t=1}^{H}  \left(\cvec{\beta}_{k,t}-\cvec{\mu}^{\mathrm{u},-}_{l_k}\right) \oslash \cvec \nu_{k,t} \\
\end{array}\right].   \hspace{1.5cm}\\
  \end{aligned}
\end{equation}
Note that for $H = 1$, i.e~a single observation, the given equation is equivalent to the RKN updates. Moreover, the given rule constitutes a unification of the batch update rule for Bayesian aggregation \cite{volpp2020bayesian} and the incremental Kalman update for our factorization of the belief state representation \cite{becker2019recurrent} detailed in Appendix A. 

\subsection{A General Definition For an  N-level MTS3}  
\label{subsec: Nlevel}
An N-level MTS3 can be defined as a family of N-state space models, $\{S_0, S_1, ..., S_{N-1}\}$. Each of the state space model $S_i$ is given by $S_i = (Z_i, A_i, O_i, f_i, h_i, H_i \Delta t, L_i)$, where $Z_i$ is the state space, $A_i$ the action space, and $O_i$ the observation space of the SSM. The parameter $H_i \Delta t$ denotes the discretization time-step and $f_i$ and $h_i$ the dynamics and observation models, respectively. Here, $l_i \in L_i$ is a task descriptor that parametrizes the dynamics model of the SSM and is held constant for a local window of $H_{i+1}$ steps. $l_i$ is a function of the latent state of SSM one level above it, i.e., $S_{i+1}$. The boundary cases can be defined as follows: for $i=0$, $H_0 = 1$. Similarly, for $i=N-1$, the latent task descriptor $L_i$ is an empty set. For all $i$, $H_i < H_{i+1}$.

Even though our experiments focus on MTS3 models with 2 hierarchies, extensive experimentation with more hierarchies can be taken as future work.

\section{MTS3 as a Hierarchical World Model}

MTS3 allows for a natural way to build world models that can deal with partial observability, non-stationarity and uncertainty in long-term predictions, properties which are critical for model-based control and planning. Furthermore, introducing several levels of latent variables, each working at a  different time scale allows us to learn world models that can make action conditional predictions/``dreams'' at multiple time scales and multiple levels of state and action abstractions. 

\subsection{Conditional Multi Time Predictions With World Model} 

Conditional multi-step ahead predictions involve estimating plausible future states of the world resulting from a sequence of actions. Our principled formalism allows for action-conditional future predictions at multiple levels of temporal abstractions. The prediction update for the sts-SSM makes prior estimates about future latent task variables conditioned on the abstract action representations. Whereas, the task conditional prediction update in the fts-SSM estimates the future prior latent states, conditioned on primitive actions and the inferred latent task priors, which are decoded to reconstruct future observations. 
For initializing the prior belief $p(\cvec z_{k,1})$ for the first time step of the time window $k$, we use the prior belief $p(\cvec z_{k-1, H+1})$ of the last time step of the time window $k-1$.


\subsection{Optimizing the Predictive Log-Likelihood} 
The training objective for the MTS3 involves maximizing the posterior predictive log-likelihood which is given below for a single trajectory, i.e., 
\begin{align}
\label{eq:objective}
     L & = \sum_{k=1}^N \sum_{t=1}^H \log p(\cvec{o}_{k,t+1}|\cvec{\beta}_{1:k-1},\cvec{\alpha}_{1:k},\cvec{w}_{k,1:t}, \cvec{a}_{k,1:t}) \nonumber \\ & = \sum_{k=1}^N \sum_{t=1}^H \log \iint p(\cvec{o}_{k,t+1}|\cvec{z}_{k,t+1})  p(\cvec{z}_{k,t+1}|\cvec{w}_{k,1:t}, \cvec a_{k,1:t}, \cvec l_k) p(\cvec l_k|\cvec{\beta}_{1:k-1},\cvec{\alpha}_{1:k})d\cvec{z}_{k,t+1} d \cvec l_k \nonumber \\
     & =  \sum_{k=1}^N \sum_{t=1}^H \log \int p(\cvec{o}_{k,t+1}|\cvec{z}_{k,t+1})  p_{\cvec l_k}(\cvec{z}_{k,t+1}|\cvec{w}_{k,1:t}, \cvec a_{k,1:t}) d\cvec{z}_{k,t+1}.  
\end{align}
The extension to multiple trajectories is straightforward and omitted to keep the notation uncluttered. Here, $\cvec{o}_{k,t+1}$ is the ground truth observations at the time step $t+1$ and time window $k$ which needs to be predicted from all (latent and abstract) observations up to time step $t$. The corresponding latent state prior belief  $p_{\cvec l_k}(\cvec{z}_{k,t+1}|\cvec{w}_{k,1:t}, \cvec a_{k,1:t})$ has a closed form solution as discussed in Section \ref{sec: fts-inf}. 
\par
We employ a Gaussian approximation of the posterior predictive log-likelihood of the form $ p(\cvec{o}_{k,t+1}|\cvec{\beta}_{1:k-1},\cvec{\alpha}_{1:k},\cvec{w}_{k,1:t}, \cvec{a}_{k,1:t}) \approx \mathcal{N}(\cvec{\mu}_{\cvec{o}_{k,t+1}},\textrm{diag}(\cvec{\sigma}_{\cvec{o}_{k,t+1}}))$ where we use the mean of the prior belief $\cvec{\mu}_{z_{k,t+1}}^-$ to decode the predictive mean, i.e, $\cvec{\mu}_{\cvec{o}_{k,t+1}} =  \textrm{dec}_{\cvec{\mu}}(\cvec{\mu}_{z_{k,t+1}}^{-})$ and the variance estimate of the prior belief to decode the observation variance, i.e., $\cvec{\sigma}_{o_{k,t+1}} = \textrm{dec}_{\sigma}(\cmat{\Sigma}_{z_{k,t+1}}^{-})$. This approximation can be motivated by a moment-matching perspective and allows for end-to-end optimization of the log-likelihood without using auxiliary objectives such as the ELBO \cite{becker2019recurrent}.

Gradients are computed using (truncated) backpropagation through time (BPTT)~\cite{werbos1990backpropagation} and clipped.  
We optimize the objective using the Adam~\cite{kingma2014adam} stochastic gradient descent optimizer with default parameters. We refer to Appendix A for more details. For training, we also initialize the prior belief $p(\cvec z_{k,1})$ with the  prior belief $p_{\cvec l_{k-1}}(\cvec z_{k-1,H+1}|\cvec w_{k-1,1:H},\cvec a_{k-1,1:H})$ from of the previous time window $k-1$. However, we cut the gradients for the fast time scale between time windows as this avoids vanishing gradients and we observed a more stable learning behaviour. Yet, the gradients can still flow between time windows for the fts-SSM via the sts-SSM.

\subsection{Imputation Based Training For Long Term Prediction \label{sec:imputation}} 

Using the given training loss results in models that are good in one-time step prediction but typically perform poorly in long-term predictions as the loss assumes that observations are always available up to time step $t$. To increase the long-term prediction performance, we can treat the long-term prediction problem as a case of the ``missing value'' problem, where the missing observations are at the future time steps. Thus, to train our model for long-term prediction, we randomly mask a fraction of observations and explicitly task the network to impute the missing observations, resulting in a strong self-supervised learning signal for long-term prediction with varying prediction horizon length. This imputation scheme is applied at both time scales, masking out single-time steps or whole time windows of length H. The imputation mask is also randomly resampled for every mini-batch.
\section{Related Work}
\textbf{Multi Time Scale World Models} One of the early works that enabled environment models at different temporal scales to be intermixed, producing temporally abstract world models was proposed by \cite{sutton1995td}. The work was limited to tabular settings but showed the importance of learning environment dynamics at multiple abstractions. However, there have been limited works that actually solve this problem at scale as discussed in \cite{lecun2022path}. A probabilistically principled formalism for these has been lacking in literature and this work is an early attempt to address this issue.
\par
\textbf{Deep State Space Models.} Deep SSMs combine the benefits of deep neural nets and SSMs by offering tractable probabilistic inference and scalability to high-dimensional and large datasets. \cite{haarnoja2016backprop, becker2019recurrent, shaj2020action} use neural network architectures based on exact inference on SSMs and perform state estimation and dynamics prediction tasks. \cite{shaj2022hidden} extend these models to modelling non-stationary dynamics. \cite{krishnan2017structured, karl2016deep, hafner2019learning} perform learning and inference on SSMs using variational approximations. However, most of these recurrent state-space models have been evaluated on very short-term prediction tasks in the range of a few milliseconds and model the dynamics at a single time scale.
\par
\textbf{Transformers} 
Recent advancements in Transformers~\cite{vaswani2017attention, radford2019language, brown2020language}, which rely on attention mechanism, have demonstrated superior performance in capturing long-range dependency compared to RNN models in several domains including time series forecasting~\cite{zhou2021informer,liu2022nonstationary} and learning world models~\cite{micheli2023transformers}. \cite{zhou2021informer,liu2022nonstationary,nie2023a} use transformer architectures based on a direct multistep loss~\cite{zeng2022transformers} and show promising results for long-term forecasting since they avoid error accumulation from autoregression. On the other hand \cite{micheli2023transformers} uses a GPT-like autoregressive version of transformers to learn world models. These deterministic models, however, do not deal with temporal abstractions and uncertainty estimation in a principled manner. Nevertheless, we think Transformers that operate at multiple timescales based on our formalism can be a promising alternative research direction.

\section{Experiments}
In this section, we evaluate our approach to a diverse set of simulated and real-world dynamical systems for long-horizon prediction tasks. Our experiments are designed to answer the following questions. (a) Can MTS3 make accurate long-term deterministic predictions (mean estimates)? (b) Can MTS3 make accurate long-term probabilistic predictions (variance estimates)? (c) How important are the modelling assumptions and training scheme? 

\subsection{Baseline Dynamics Models}
While a full description of our baselines can be found in Appendix E, a brief description of them is given here: (a) \textbf{RNNs} - We compare our method to two widely used recurrent neural network architectures, LSTMs~\citep{lstm} and GRUs~\citep{gru}. 
(b) \textbf{RSSMs} - Among several RSSMs from the literature, we chose RKN~\citep{becker2019recurrent} and HiP-RSSM~\citep{shaj2022hidden} as these have shown excellent performance for dynamics learning for short-term predictions and rely on exact inference as in our case. (c) \textbf{Transformers} - We also compare with two state-of-the-art Transformer~\cite{vaswani2017attention} variants. The first variant (AR-Transformer) relies on a GPT-like autoregressive prediction~\cite{radford2019language,brown2020language}. Whereas the second variant (Multi-Transformer) uses direct multi-step loss~\cite{zeng2022transformers} from recent literature on long horizon time-series forecasting \cite{zhou2021informer,liu2022nonstationary,nie2023a}. Here, multistep ahead predictions are performed using a single shot given the action sequences.
\subsection{Environments and Datasets}
We experiment with three broad datasets. While full descriptions of these datasets, dataset creation procedure, and overall statistics are given in Appendix D, a brief description of them is as follows.
(a) \textbf{D4RL Datasets} - We use a set of 3 different environments/agents from D4RL dataset~\cite{fu2020d4rl}, which includes the HalfCheetah, Medium Maze and Franka Kitchen environment. Each of these was chosen because of their distinct properties like sub-optimal trajectories (HalfCheetah), realistic domains / human demonstrations (Kitchen), multi-task trajectories, non-markovian collection policies (Kitchen and Maze) and availability of long horizon episodes (all three). (b) \textbf{Manipulation Datasets} - We use 2 datasets collected from a real excavator arm and a Panda robot. The highly non-linear non-markovian dynamics due to hydraulic actuators in the former and non-stationary dynamics owing to different payloads in the latter make them challenging benchmarks. Furthermore, accurate modelling of the dynamics of these complex systems is important since learning control policies for automation directly on large excavators is economically infeasible and potentially hazardous. (c) \textbf{Mobile Robotics Dataset} - We set up a simulated four-wheeled mobile robot traversing a highly uneven terrain of varying steepness generated by a mix of sinusoidal functions. This problem is challenging due to the highly non-linear dynamics involving wheel-terrain interactions and non-stationary dynamics introduced by varying steepness levels. In all datasets, we only use information about agent/object positions and we mask out velocities to create a partially observable setting.

\begin{figure}[t]
\begin{subfigure}[b]{\textwidth}
\hspace*{-1.2cm}
         \includegraphics[width=1.15\linewidth]{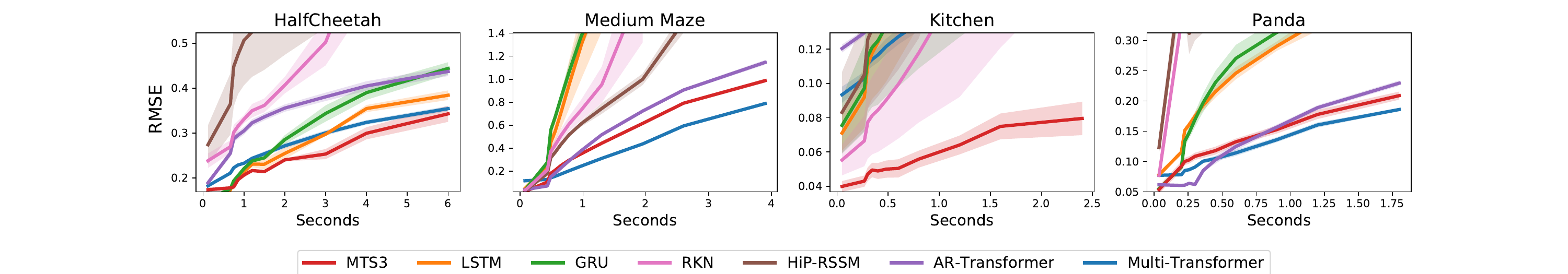}
     \caption{}
     \label{fig:m}
     \end{subfigure}
     \\
\begin{subfigure}[b]{0.56\textwidth}
\hspace*{-0.5cm}
         \includegraphics[width=\linewidth]{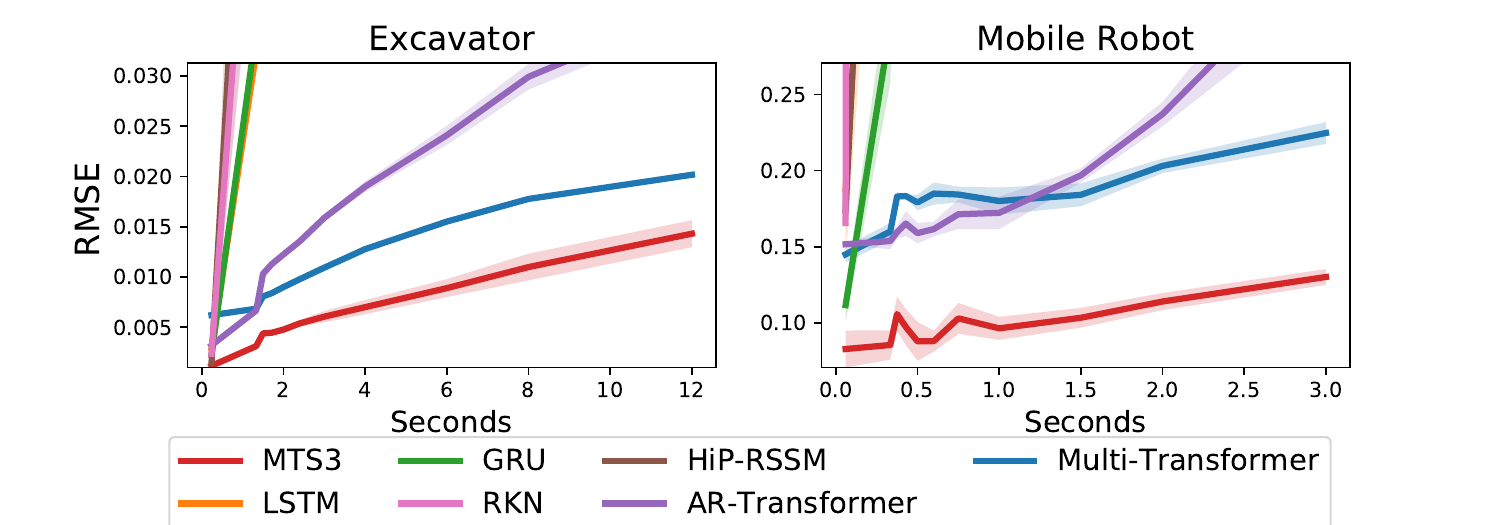}
     \caption{}
     \label{fig:m}
     \end{subfigure}
     \hfill
\begin{subfigure}[b]{0.56\textwidth}
\hspace*{-1.3cm}
         \includegraphics[width=\linewidth]{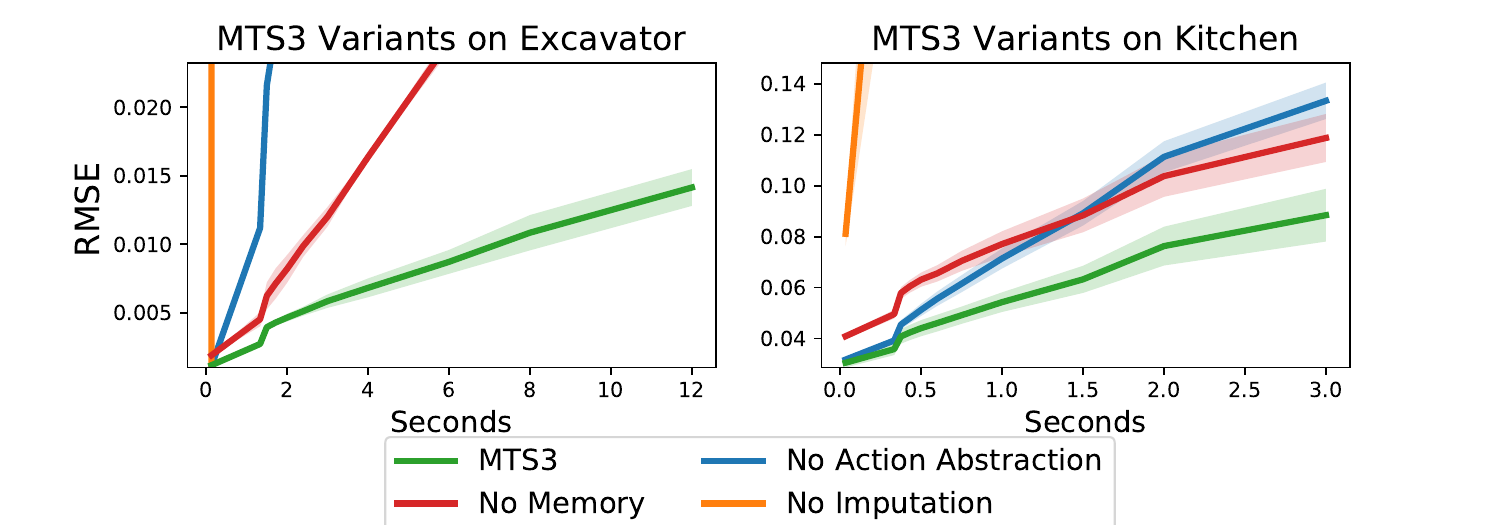}
     \caption{}
     \label{fig:abl}
     \end{subfigure}
\caption{(a) and (b) Comparison with baselines in terms of RMSE for long horizon predictions (in seconds) as discussed in Section \ref{subsec:mean}. (c) Ablation with different MTS3 variants is discussed in \ref{subsec:ablation}.}
\label{fig:exps}
\end{figure}

\subsection{Can MTS3 make accurate long-term deterministic predictions (mean estimates)?}
\label{subsec:mean}
Here we evaluate the quality of the mean estimates for long-term prediction using our approach. The results are reported in terms of RMSE in Figure \ref{fig:exps}. We see that MTS3 gives consistently good long-term action conditional future predictions on all 6 datasets. Deep Kalman models~\cite{becker2019recurrent,shaj2022hidden} which operate on a single time scale fail to give meaningful mean estimates beyond a few milliseconds. Similarly, widely used RNN baselines~\cite{lstm,gru} which form the backbone of several world models~\cite{ha2018world,hafner2019learning} give poor action conditional predictions over long horizons. AR-Transformers also fail possibly due to error accumulation caused by the autoregression. However, Multi-Transformers are a strong baseline that outperforms MTS3 in the Medium Maze and Panda dataset by a small margin. However, on more complex tasks like the Kitchen task, which requires modelling multi-object, multi-task interactions~\cite{gupta2019relay}, MTS3 is the only model that gives meaningful long horizon predictions. A visualization of the predicted trajectories vs. ground truth is given in Appendix C.   

\subsection{Can MTS3 make accurate long-term probabilistic predictions (variance estimates)?}
Next, we examine the question of whether the principled probabilistic inference translates to accurate uncertainty quantification during long-horizon predictions. We trained all the baselines with a negative log-likelihood loss and used the same as a metric to quantify the quality of uncertainty estimates. As seen in table \ref{table:nll}, MTS3 gives the most accurate uncertainty estimates in all datasets except Medium Maze, where it is outperformed by Multi-Transformer. Also, notably, AR-Transformers and deep Kalman models fail to learn any meaningful uncertainty representation when it comes to long-term predictions.
\begin{table}[h!]
    \centering
    \resizebox{\columnwidth}{!}{
    \begin{tabular}{|c|c|c|c|c|c|c|c|c|c|c|}
        \hline
        {} & \textbf{Prediction} & \multicolumn{7}{c|}{\textbf{Algorithm}} \\ \cline{3-9} 
        & \textbf{Horizon} & \textbf{MTS3} & \textbf{Multi-Trans} & \textbf{AR-Trans} & \textbf{LSTM} & \textbf{GRU} & \textbf{RKN} & \textbf{HiP-RSSM} \\ \hline
        \textbf{Half Cheetah} & 6 s & $\mathbf{-2.80 \pm 0.30}$ & $0.25 \pm 0.05$ & \xmark
        & $7.34 \pm 0.06$ & $7.49 \pm 0.04$ & \xmark & \xmark \\ \cline{2-9}
        \hline
        \textbf{Kitchen} & 2.5 s & $\mathbf{-25.74 \pm 0.12}$ & $-7.3 \pm 0.2$ & \xmark & $32.45 \pm 1.64$ & $32.72 \pm 0.65$ & \xmark & \xmark \\ \cline{2-9}
        \hline
        \textbf{Medium Maze} & 4 s & $-0.21 \pm 0.022$ & $\mathbf{-0.88 \pm 0.02}$ & \xmark & $4.03 \pm 0.32$ & $7.76 \pm 0.07$ & \xmark & \xmark \\ \cline{2-9}
        \hline
        \textbf{Panda} & 1.8 s & $\mathbf{2.79 \pm 0.32}$ & $3.77 \pm 0.33$ & \xmark & $7.94 \pm 0.39$ & $7.91 \pm 0.23$ & \xmark & \xmark \\ \cline{2-9}
        \hline
        \textbf{Hydraulic} & 12 s & $\mathbf{-2.64 \pm 0.12}$ & $-2.46 \pm 0.03$ & \xmark & $7.35 \pm 0.061$ & $7.35 \pm 0.06$ & \xmark & \xmark \\ \cline{2-9}
        \hline
        \textbf{Mobile Robot} & 3 s & $\mathbf{-6.47 \pm 0.71}$ & $-5.17 \pm 0.23$ & \xmark & $11.27 \pm 2.3$ & $14.55 \pm 5.6$ & \xmark & \xmark \\ \cline{2-9}
        \hline
    \end{tabular}
    }
    \caption{Comparison in terms of Negative Log Likelihood (NLL) for long horizon predictions (in seconds). Here bold numbers indicate the top methods and \xmark~denotes very high/nan values resulting from the highly divergent mean/variance long-term predictions.}
    \label{table:nll}
\end{table}

\subsection{How important are the modelling assumptions and training scheme?}
\label{subsec:ablation}
Now, we look at three important modelling and training design choices: (i) splitting the latent states to include an unobservable ``memory'' part using observation model $h^{sts}=h^{fts}=\cmat H = [\cmat I, \cmat 0]$ as discussed in Sections \ref{sts-inf} and \ref{sec: fts-inf}, (ii) action abstractions discussed in Section \ref{sts}, (iii) training by imputation. 


To analyze the importance of the memory component, we derived and implemented an MTS3 variant with an observation model of $h^{sts}=h^{fts}=\cmat I$ and a pure diagonal matrix representation for the covariance matrices. As seen in Figure \ref{fig:abl}, this results in worse long-term predictions, suggesting that splitting the latent states in its observable and unobservable part in MTS3 is critical for learning models of non-markovian dynamical systems. Regarding (ii), we further devised another variant where MTS3 only had access to observations, primitive actions and observation abstractions, but no action abstractions. As seen in our ablation studies, using the action abstraction is crucial for long-horizon predictions.

Our final ablation (iii) shows the importance of an imputation-based training scheme discussed in Section \ref{sec:imputation}. As seen in Figure \ref{fig:abl} when trained for 1 step ahead predictions without imputation, MTS3 performs significantly worse for long-term prediction suggesting the importance of this training regime.


\subsection{What is the role of the discretization step $H.\Delta t$?}
\label{subsec:abl_h}
\begin{figure}
\centering
\begin{minipage}{.29\textwidth}
  \centering
  \includegraphics[width=.98\linewidth]{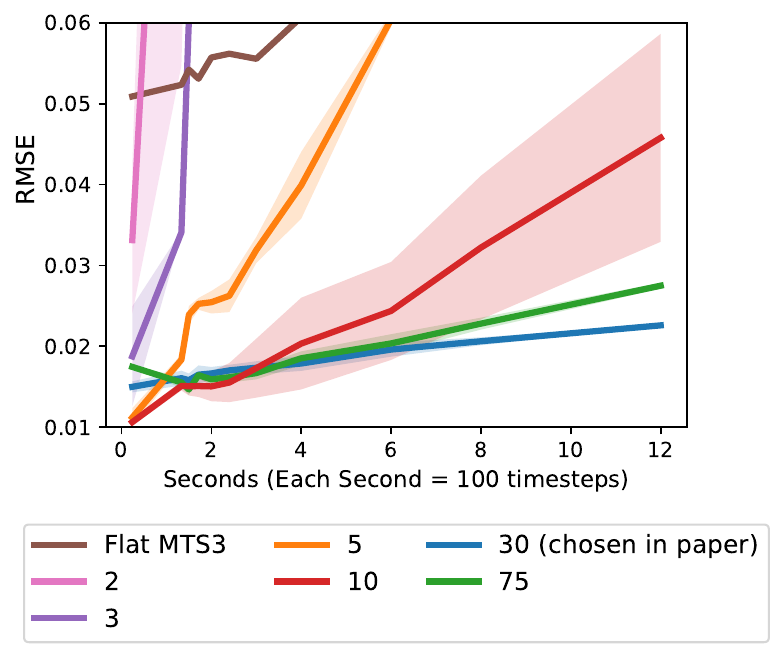}
  \subcaption{}
  \label{fig:abl1}
\end{minipage}%
\begin{minipage}{.71\textwidth}
  \centering
  \includegraphics[width=.98\linewidth]{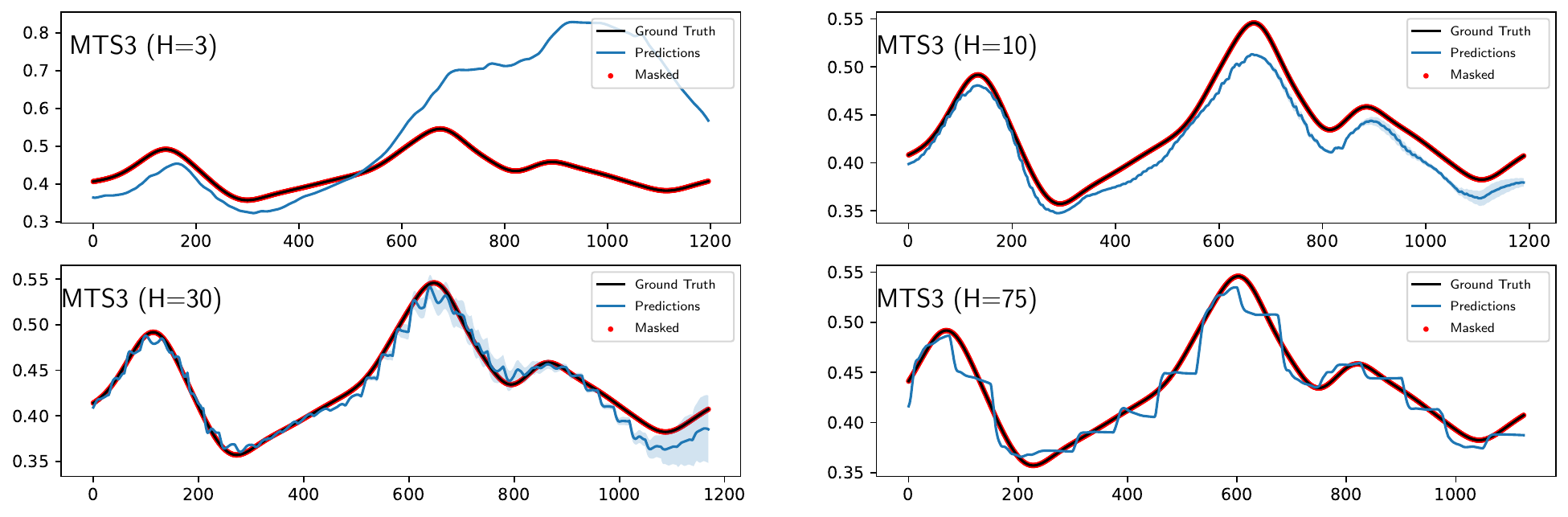}
  \subcaption{}
  \label{fig:abl2}
\end{minipage}
\caption{Ablation on discretization step $H. \Delta t$ (a) The long-term prediction results in terms of RMSE, with different $H$ values as discussed in Section \ref{subsec:abl_h} on the hydraulics dataset. (b) The predictions by MTS3 variants with different values of timescale parameter $H.\Delta t$ on a trajectory picked from the hydraulics excavator dataset. The top figures are for $H=3$ and 
$H=10$. Bottom figures are for $H=30$ and $H=75$. Note that the results reported in the paper are with $H=30$.}
\end{figure}

Finally, we perform ablation for different values of $H.\Delta t$, which controls the time scale of the task dynamics. The results reported are for the hydraulics dataset. The higher the value of H, the slower the timescale of the task dynamics relative to the state dynamics. As seen in Figure \ref{fig:abl1}, smaller values of $H$ (2,3,5 and 10) give significantly worse performance. Very large values of $H$ (like 75) also result in degradation of performance. To further get an intuitive understanding of the MTS3's behaviour under different timescales, we plot the predictions given by MTS3 for different values of $H$ on a trajectory handpicked from the hydraulics excavator dataset. As seen in Figure \ref{fig:abl2}, for large values of $H$ like 30 and 75, we notice that the slow-changing task dynamics "reconfigures" the fast dynamics every 30 and 75-step window respectively, by conditioning the lower level dynamics with the newly updated task prior. This effect is noticeable as periodic jumps or discontinuities in the predictions, occurring at 30 and 75-step intervals. Also, for a very large $H$ like 75, the fast time scale ssm has to make many more steps in a longer window resulting in error accumulation and poor predictions. 
\label{subsec:ablation}

\section{Conclusion and Future Work}
In this work, we introduce MTS3, a probabilistic formalism for learning the dynamics of complex environments at multiple time scales. By modelling the dynamics of the world at multiple levels of temporal abstraction we capture both the slow-changing long-term trends and fast-changing short-term trends in data, leading to highly accurate predictions spanning several seconds into the future. Our experiments demonstrate that simple linear models with principled modelling assumptions can compete with large transformer model variants that require several times more parameters. Furthermore, our inference scheme also allows for principled uncertainty propagation over long horizons across multiple time scales which capture the stochastic nature of environments. We believe our formalism can benefit multiple future applications including hierarchical planning/control. We discuss the limitations and broader impacts of our work in Appendix F and G.

\newpage
\section{Acknowledgement}
We thank the anonymous reviewers for the valuable remarks and discussions which greatly improved the quality of this paper. This work was supported by funding from the pilot program Core Informatics of the Helmholtz Association (HGF). The authors acknowledge support by the state of Baden-Württemberg through bwHPC, as well as the HoreKa supercomputer funded by the Ministry of Science, Research and the Arts Baden-Württemberg and by the German Federal Ministry of Education and Research.

\bibliographystyle{plainnat}
\bibliography{neurips_2023}
\newpage
\appendix
\section{Implementation Details}

 \begin{figure*}[h]
\begin{center}
\includegraphics[scale=0.12]{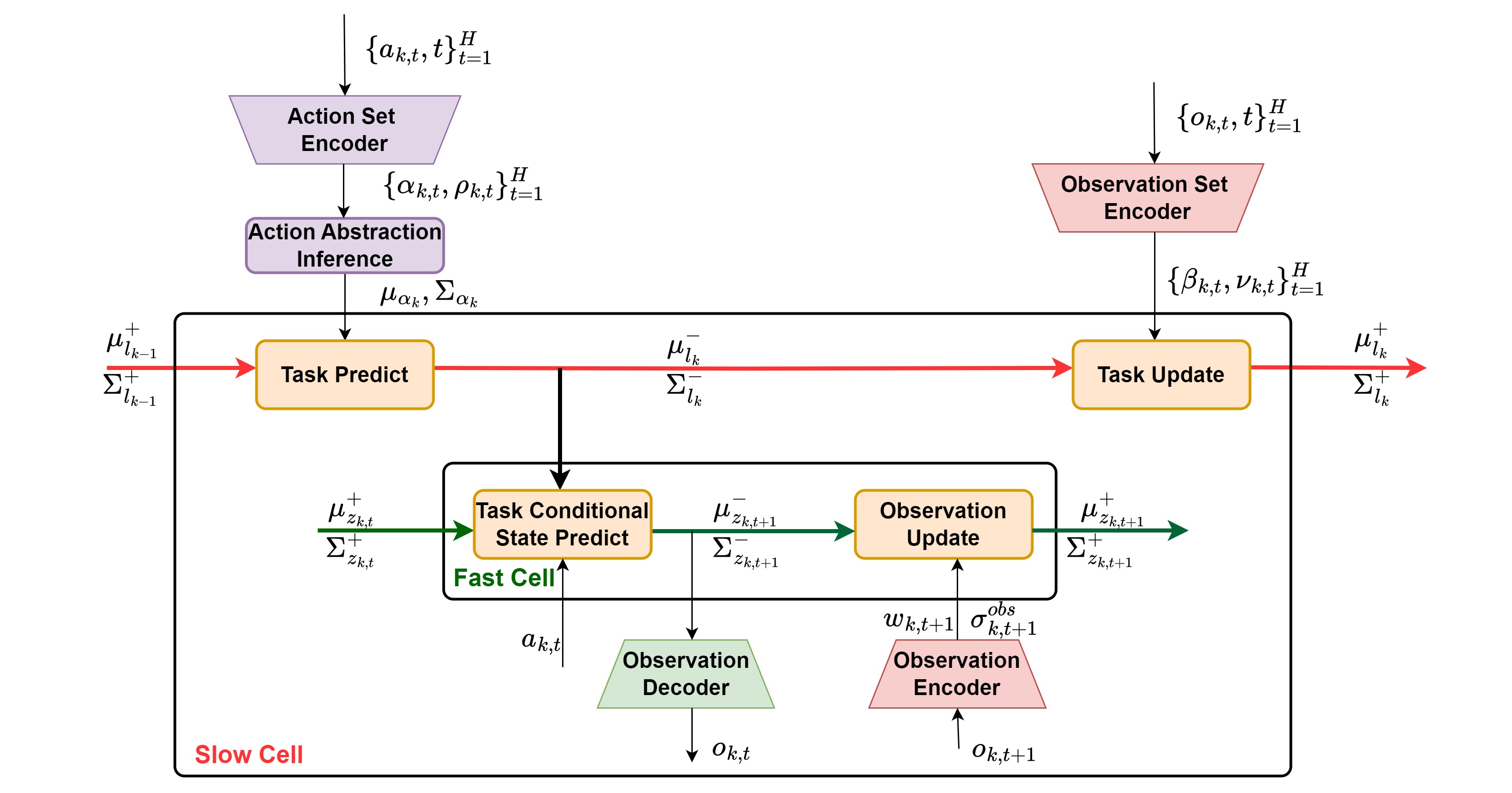}
\end{center}
\caption{Schematic of a 2-Level MTS3 Architecture. Inference in MTS3 takes place via closed-form equations derived using exact inference, spread across two-time scales. For the fast time scale (fts) SSM, these include the task conditional state predict and observation update stages as discussed in Section 3.2 of the main paper. Whereas, for the slow time scale (sts) SSM, these include the task prediction and task update stages which are described in Section 3.3.  } 
 \label{fig:schematic}
\end{figure*}

\newpage 
\subsection{Inference In Slow Time Scale SSM}

\subsubsection{Inferring Action Abstraction (sts-SSM)}
\begin{wrapfigure}[11]{r}{.31\linewidth}
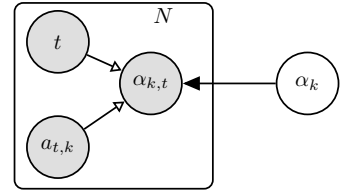

\begin{subfigure}[b]{0.31\textwidth}
   \centering\begin{adjustbox}{width=\textwidth}
         \tikzActAb
     \end{adjustbox}
     \end{subfigure}
\caption{Generative model for the abstract action $\alpha_k$. The hollow arrows are deterministic transformations leading to implicit distribution $\alpha_{k,t}$ using an action set encoder.}
\label{fig:pgmactabs}
\end{wrapfigure}
Given a set of encoded primitive actions and their corresponding variances $ \{ \cvec{\alpha_{k,t}}, \cvec{ \rho_{k,t}} \}_{t=1}^H $, using the prior and observation model assumptions in Section 3.1.2 of main paper, we infer the latent abstract action $p(\cvec \alpha_{k}|\cvec \alpha_{k,1:H}) = \mathcal{N}(\cvec \mu_{ \alpha_{k}}, \cvec \Sigma_{\alpha_k})  = \mathcal{N}(\cvec \mu_{ \alpha_{k}}, \textrm{diag}\cvec(\sigma_{\alpha_k}))$ as a Bayesian aggregation~\cite{volpp2020bayesian} of these using the following closed-form equations:

\begin{align*}
\cvec{\sigma}_{\alpha_{k}} & = \left( \left( \cvec{\sigma}_{0} \right)^\ominus + \sum_{n=1}^N \left(\left(\cvec{\rho_{k,t}}\right)^\ominus\right) \right)^\ominus, \\
\cvec{\mu}_{\alpha_{k}} & = \cvec{\mu}_{0} + \cvec{\sigma}_{\alpha_{k}} \odot \sum_{n=1}^N \left(\cvec{\alpha_{k,t}} - \cvec{\mu}_{0} \right) \oslash \cvec{\rho_{k,t}}
\end{align*}

Here, $\ominus$, $\odot$ and $\oslash$ denote element-wise inversion, product, and division, respectively. The update equation is coded as the ``abstract action inference'' neural network layer as shown in Figure \ref{fig:schematic}. 

\subsubsection{Task Prediction (sts-SSM)}

The goal of this step is to update the prior marginal over the latent task variable $\cvec{l}_k$, $p(\cvec{l}_k|\cvec{\beta}_{1:k-1},\cvec{\alpha}_{1:k})$, given the posterior beliefs from the time window $k-1$ and abstract action $\cvec \alpha_{k}$.

Using the linear dynamics model assumptions from Section 3.3, we can use the following closed-form update equations to compute,
$p(\cvec{l}_k|\cvec{\beta}_{1:k-1},\cvec{\alpha}_{1:k}) = \mathcal{N}(\cvec{\mu}_{l_{k}}^-,\cvec{\Sigma}_{l_{k}}^-)$, where
\begin{equation}
\label{eq:task-predict}
\begin{aligned}[c]
\cvec{\mu}_{l_{k}}^- &=\cmat{X}\cvec{\mu}_{l_{k-1}}^+ + \cmat{Y}\cvec{\alpha}_{k}\\
   \cvec{\Sigma}_{l_{k}}^- &= \cmat{X}\cvec{\Sigma}_{l_{k-1}}^+\cmat{X}^T + \cmat{Y}\cvec{\Sigma}_{\alpha_{k}}\cmat{Y}^T + \cmat S.
  \end{aligned}
\end{equation}

These closed-form equations are coded as the ``task predict'' neural net layer as shown in Figure \ref{fig:schematic}.

\subsubsection{Task Update (sts-SSM)}

In this stage, we update the prior over $l_k$ using an abstract observation set $\{\cvec \beta_{k,t}\}_{t=1}^H$, to obtain the latent task the posterior $\mathcal{N}(\cvec{\mu}_{z_{k,t}}^+,\cmat{\Sigma}_{z_{k,t}}^+) = \mathcal{N}(\left[\begin{array}{c}\cvec{\mu}_{t}^{u+} \\
\cvec{\mu}_{t}^{l+} \end{array}\right],\left[\begin{array}{cc}\cvec \Sigma_t^u & \cvec \Sigma^{s}_t \\ \cvec \Sigma^{s}_t & \cvec \Sigma^{l}_t  \end{array}\right]^+)$, with $\cvec \Sigma_{l_k}^u = \textrm{diag}(\cvec \sigma_{l_k}^u), \; \cvec \Sigma_{l_k}^l = \textrm{diag}(\cvec \sigma_{l_k}^l)$ \textrm{ and } $\cvec \Sigma_{l_k}^{s} = \textrm{diag}(\cvec \sigma_{l_k}^s)$.

To do so we first invert the prior covariance matrix $\left[\begin{array}{cc}\cvec \Sigma_{l_k}^u & \cvec \Sigma_{l_k}^{s} \\ \cvec \Sigma_{l_k}^{s} & \cvec \Sigma_{l_k}^{l}  \end{array}\right]^+$ to the precision matrix $\left[\begin{array}{cc}\cvec \lambda_{l_k}^u & \cvec \lambda_{l_k}^{s} \\ \cvec \lambda_{l_k}^{s} & \cvec \lambda_{l_k}^{l}  \end{array}\right]^+$ for permutation invariant parallel processing. The posterior precision is then computed using scalar operations are follows, where only $\cvec \lambda_{l_{k}}^{u}$ is changed by 
\begin{align}\cvec \lambda_{l_{k}}^{u+} = \cvec \lambda_{l_{k}}^{u-} + \sum_{t=1}^H \cvec 1 \oslash \cvec \nu_{k,t}\end{align}
 while $\cvec \lambda_{l_{k}}^{l+} = \cvec \lambda_{l_{k}}^{l-}$ and $\cvec \lambda_{l_{k}}^{s+} = \cvec \lambda_{l_{k}}^{s-}$ remain constant. The operator $\oslash$ denotes the element-wise division. The posterior precision is inverted back to the posterior covariance vectors $\cvec \sigma_{l_{k}}^{u+}$, $\cvec \sigma_{l_{k}}^{l+}$ and $\cvec \sigma_{l_{k}}^{s+}$. Now, the posterior mean $\cvec{\mu}_{l,k}^{+}$ can be obtained from the prior mean $\cvec{\mu}_{l,k}^{-}$ as
 \begin{equation}
  \begin{aligned}[c]
   \cvec{\mu}_{l,k}^{+}=\cvec{\mu}_{l,k}^{-} + \left[ \begin{array}{l}
\cvec{\sigma}_{l_k}^{u+} \\
\cvec{\sigma}_{l_k}^{s+} \\
\end{array}\right]  \odot \left[ \begin{array}{l}
\sum_{t=1}^{H}  \left(\cvec{\beta}_{k,t}-\cvec{\mu}^{\mathrm{u},-}_{l_k}\right) \oslash \cvec \nu_{k,t} \\
\sum_{t=1}^{H}  \left(\cvec{\beta}_{k,t}-\cvec{\mu}^{\mathrm{u},-}_{l_k}\right) \oslash \cvec \nu_{k,t} \\
\end{array}\right].   \hspace{1.5cm}\\
  \end{aligned}
\end{equation}
 \begin{figure*}[h]
\begin{center}
\includegraphics[scale=0.33]{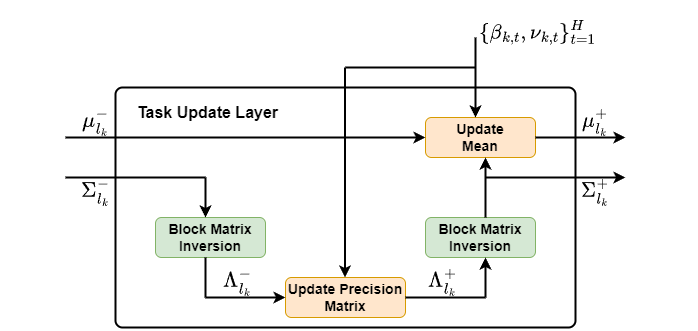}
\end{center}
\caption{Implementation of task update layer which performs posterior latent task inference in the sts-SSM.} 
 \label{fig:schematicTaskU}
\end{figure*}
\\
The inversion between the covariance matrix and precision matrix can be done via scalar operations leveraging block diagonal structure as derived in Appendix B. Figure \ref{fig:schematicTaskU} shows the schematic of the task update layer.
\subsection{Inference In Fast Time Scale SSM}

The inference in fts-SSM for a time-window $k$ involves two stages as illustrated in Figure \ref{fig: schematic}, calculating the prior and posterior over the latent state variable $z_t$. To keep the notation uncluttered, we will also omit the time-window index $k$ whenever the context is clear as in section 3.2.

\subsubsection{Task Conditional State Prediction (fts-SSM)}

Following the assumptions of a task conditional linear dynamics as in Section 3.2 of the main paper, we obtain the prior marginal for $p(\cvec{z}_{k,t}|\cvec{w}^k_{1:t-1},\cvec{a}^k_{1:t-1},\cvec{\beta}_{1:k-1},\cvec{\alpha}_{1:k-1}) = \mathcal{N}(\cvec{\mu}_{z_{k,t}}^-,\cmat{\Sigma}_{z_{k,t}}^-)$ in closed form, where
\begin{equation}
\label{eq:cond-predict}
\centering
\begin{aligned}
\cvec{\mu}_{z_{k,t}}^- &=\cmat{A}\cvec{\mu}_{z_{k,t-1}}^- + \cmat{B}\cvec{a}_{k,t-1} + \cmat{C}\cvec{\mu}_{l_{k}}^-, \\
   \cmat{\Sigma}_{k,t}^- &= \cmat{A}\cmat{\Sigma}_{k,t-1}^+\cmat{A}^T + \cmat{C}\cvec{\Sigma}_{l_{k}}^-\cmat{C}^T + \cmat{Q}.
  \end{aligned}
\end{equation}

\subsubsection{Observation Update (fts-SSM)}
In this stage, we compute the posterior belief $p(\cvec{z}_{k,t}|\cvec{w}^k_{1:t},\cvec{a}^k_{1:t},\cvec{\beta}_{1:k},\cvec{\alpha}_{1:k-1}) = \mathcal{N}(\cvec{\mu}_{z_{k,t}}^-,\cmat{\Sigma}_{z_{k,t}}^-)$.  using the same closed-form update as in  \cite{becker2019recurrent}. The choice of the special observation model splits the state into two parts, an upper $\boldsymbol{z}_t^\textrm{u}$ and a lower part $\boldsymbol{z}_t^\textrm{l}$, resulting in the posterior belief $\mathcal{N}(\cvec{\mu}_{z_{k,t}}^-,\cmat{\Sigma}_{z_{k,t}}^-) = \mathcal{N}(\left[\begin{array}{c}\cvec{\mu}_{t}^{u+} \\
\cvec{\mu}_{t}^{l+} \end{array}\right], \left[\begin{array}{cc}\cvec \Sigma_t^u & \cvec \Sigma^{s}_t \\ \cvec \Sigma^{s}_t & \cvec \Sigma^{l}_t  \end{array}\right]^+)$, with $\cvec \Sigma_t^u = \textrm{diag}(\cvec \sigma_t^s), \; \cvec \Sigma_t^l = \textrm{diag}(\cvec \sigma_t^l)$ and $\cvec \Sigma_t^{s} = \textrm{diag}(\cvec \sigma_{t}^s)$. Thus, the factorization allows for only the diagonal and one off-diagonal vector of the covariance to be computed and simplifies the calculation of the mean and posterior to simple scalar operations. 

The closed-form equations for the mean can be expressed as the following scalar equations,
\begin{align*}
\boldsymbol{z}_t^+ = \boldsymbol{z}_t^- +
\left[\begin{array}{c} \boldsymbol{\sigma}^\mathrm{u,-}_t \\
\boldsymbol{\sigma}^\mathrm{l,-}_t \end{array}\right]
\odot
\left[\begin{array}{c}\boldsymbol{w}_t - \boldsymbol{z}^{\mathrm{u},-}_t \\
\boldsymbol{w}_t - \boldsymbol{z}^{\mathrm{u},-}_t  \end{array}\right]
\oslash 
\left[\begin{array}{c}  \boldsymbol{\sigma}_t^{\mathrm{u},-} + \boldsymbol{\sigma}_t^\mathrm{obs} 
 \\  \boldsymbol{\sigma}_t^{\mathrm{u},-} + \boldsymbol{\sigma}_t^\mathrm{obs} \end{array}\right],
\end{align*}

The corresponding equations for the variance update can be expressed as the following scalar operations,
\begin{align*}
\boldsymbol{\sigma}^{\mathrm{u},+}_t &= \boldsymbol{\sigma}^{\mathrm{u},-}_t \odot \boldsymbol{\sigma}^{\mathrm{u},-}_t \oslash \left( \boldsymbol{\sigma}_t^{\mathrm{u},-} + \boldsymbol{\sigma}_t^\mathrm{obs} \right),  \\
\boldsymbol{\sigma}^{\mathrm{s},+}_t &= \boldsymbol{\sigma}^{\mathrm{u},-}_t \odot \boldsymbol{\sigma}^{\mathrm{s},-}_t \oslash \left( \boldsymbol{\sigma}_t^{\mathrm{u},-} + \boldsymbol{\sigma}_t^\mathrm{obs} \right), \\
\boldsymbol{\sigma}^{\mathrm{l},+}_t &= \boldsymbol{\sigma}^{\mathrm{l}, -}_t - \boldsymbol{\sigma}^{\mathrm{s},-}_t \odot \boldsymbol{\sigma}^{\mathrm{s},-}_t \oslash \left( \boldsymbol{\sigma}_t^{\mathrm{u},-} + \boldsymbol{\sigma}_t^\mathrm{obs} \right),
\end{align*},
where $\odot$ denotes the elementwise vector product and  $\oslash$ denotes an elementwise vector division. 

\subsection{Modelling Assumptions}

\subsubsection{Control Model} To achieve action conditioning within the recurrent cell of fts-SMM, we include a control model $b(a_{k,t})$ in addition to the linear transition model $A_t$. $b(a_{k,t}) =f(a_{k,t})$, where $f(.)$ can be any non-linear function approximator.  We use a multi-layer neural network regressor with ReLU activations~\cite{shaj2020action}. 

However, unlike the fts-SSM where actions are assumed to be known and subjected to no noise, in the sts-SSM, the abstract action is an inferred latent variable with an associated uncertainty estimate. Hence we use a linear control model $Y$, for principled uncertainty propagation. 

\subsubsection{Transition Noise}
We assume the covariance of the transition noise $Q$ and $S$ in both timescales to be diagonal. The noise is learned and is independent of the latent state.

\subsection{Training}
\subsubsection{Training Objective Derivation} We further expand on the training objective in Section 4.2 here. The training objective for the MTS3 involves maximizing the posterior predictive log-likelihood which for a single trajectory, can be derived as, 
\begin{align}
\label{eq:objective}
     L & = \sum_{k=1}^N \sum_{t=1}^H \log p(\cvec{o}_{k,t+1}|\cvec{\beta}_{1:k-1},\cvec{\alpha}_{1:k-1},\cvec{w}_{k,1:t}, \cvec{a}_{k,1:t}) \nonumber \\ & = \sum_{k=1}^N \sum_{t=1}^H \log \iint p(\cvec{o}_{k,t+1}|\cvec{z}_{k,t+1})  p(\cvec{z}_{k,t+1}|\cvec{w}_{k,1:t}, \cvec a_{k,1:t}, \cvec l_k) p(\cvec l_k|\cvec{\beta}_{1:k-1},\cvec{\alpha}_{1:k-1})d\cvec{z}_{k,t+1} d \cvec l_k \nonumber \\
     & =  \sum_{k=1}^N \sum_{t=1}^H \log \int p(\cvec{o}_{k,t+1}|\cvec{z}_{k,t+1})  p_{\cvec l_k}(\cvec{z}_{k,t+1}|\cvec{w}_{k,1:t}, \cvec a_{k,1:t}) d\cvec{z}_{k,t+1}.  
\end{align}

The extension to multiple trajectories is straightforward. The approximation to the objective is done based on a moment-matching perspective as discussed in Section 4.2 of the main paper.

\subsubsection{Initialization}
We initialize the states $\cvec l_1$ and $\cvec z_{1,1}$ at both timescales for the first-time window $k=1$ with an all zeros vector and corresponding  covariance matrices as $\cmat \Sigma_{l_1} = \cmat \Sigma_{z_{1,1}} = 10 \cdot \cmat I$. For subsequent windows,  the prior belief $p(\cvec z_{k,1})$ for the first time step of time window $k$, is initialized using the posterior belief $p_{\cvec l_{k-1}}(\cvec z_{k-1,H}|\cvec w_{k-1,1:H},\cvec a_{k-1,1:H})$ of the last time step of  time window $k-1$.

 It is also crucial to correctly initialize the transition matrix at both time scales so that the transition does not yield an unstable system. Initially, the transition model should focus on copying the encoder output so that the encoder can learn how to extract good features if observations are available and useful. We initialize the diagonal elements of the transition matrix at both timescales with 1 and the off-diagonal elements with 0.2, while the rest of the elements are set to 0, a choice inspired from \cite{becker2019recurrent}.

 \subsubsection{Learnable Parameters}

 The learnable parameters in the computation graph are as follows:
 
 \paragraph{Fast Time Scale SSM:} The linear transition model A, the non-linear control factor b, the linear latent task transformation model C, the transition noise Q, along with the observation encoder and the output decoder.

 \paragraph{Slow Time Scale SSM:} The linear transition model X, the linear control model Y, the transition noise S, along with the observation set encoder and the action set encoder.

\newpage 
\section{Proofs and Derivations}

\begin{wrapfigure}[7]{r}{.29\linewidth}
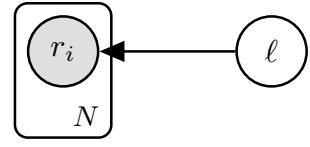

\begin{subfigure}[b]{0.28\textwidth}
   \centering \begin{adjustbox}{width=\textwidth}
            \tikzTaskinfer
     \end{adjustbox}
\end{subfigure}
\caption{Graphical Model For Bayesian conditioning with $N$ observations.}
\label{fig:taskinf}
\end{wrapfigure}

In the following sections vectors are denoted by a lowercase letter in bold, such as "$\cvec{\mathrm{v}}$", while Matrices as an uppercase letter in bold, such as "$\cmat{M}$". $\cmat{I}$ denotes identity matrix and $\cmat{0}$ represents a matrix filled with zeros. For any matrix $\cmat{M}$, $\cvec{m}$ denotes the corresponding vector of diagonal entries. Also, $\odot$ denotes the elementwise vector product and  $\oslash$ denotes an elementwise vector division.

\subsection{Bayesian Conditioning As Permutation Invariant Set Operations}

\begin{GU}[Bayesian Conditioning]
\label{gu1} Consider the graphical model given in Figure \ref{fig:taskinf}, where a set of N conditionally i.i.d observations $\cvec{\Bar{r}} = \{\cvec{r}_i\}_{i=1}^N$ are generated by a latent variable $\cvec{l}$ and the observation model $p(\cvec{r}_i|\cvec{l}) = \mathcal{N}\left(\cvec{r}_i \mid \cmat{H} \cvec{l}, \textrm{diag}(\cmat{\sigma}_i^{obs})\right)$. Assuming an observation model $\cmat{H}=[\cmat{I},\cmat{0}]$, the mean ($\cvec{\mu}$) and precision matrix ($\cvec{\Lambda}$) of the posterior over the latent variable $\cvec{l}$, $p(\cvec{l}|\cvec{\Bar{r}}) = \mathcal{N}\left( \cvec{\mu}_l^{+}, \cmat{\Sigma}_l^{+} \right) = \mathcal{N}\left( \cvec{\mu}_l^{+}, (\cmat{\Lambda}_l^{+})^{-1} \right)$, given the prior $p_0(\cvec{l}) = \mathcal{N}\left( \cvec{\mu}_l^{-}, \cmat{\Sigma}_l^{-} \right) = \mathcal{N}\left( \cvec{\mu}_l^{-}, (\cmat{\Lambda}_l^{-})^{-1} \right)$ have the following permutation invariant closed form updates.

\begin{equation}
\begin{aligned}
\cmat{\Lambda}_l^{+} & = \cmat{\Lambda}_l^{-} + \left[ \begin{array}{ll}
\textrm{diag}(\sum_{i=1}^n\frac{1}{\cvec{\sigma}_i^{obs}}), &\cvec{0 }  \\
\cvec{0 },& \cvec{0} \\
\end{array}\right] \\
   \cvec{\mu}_l^{+}&=\cvec{\mu}_l^{-} + \left[ \begin{array}{l}
\cvec{\sigma}_l^{u+} \\
\cvec{\sigma}_l^{s+} \\
\end{array}\right]  \odot \left[ \begin{array}{l}
\sum_{i=1}^{N}  \left(\cvec{r}_i-\cvec{\mu}^{\mathrm{u},-}_l\right) \odot \frac{1}{\cvec{\sigma}_i^{obs}} \\
\sum_{i=1}^{N}  \left(\cvec{r}_i-\cvec{\mu}^{\mathrm{u},-}_l\right) \odot \frac{1}{\cvec{\sigma}_i^{obs}
} \\
\end{array}\right]   \hspace{1.5cm}\\
  \end{aligned}
\label{eq: ba2}
\end{equation}
\end{GU}

Note that $\cvec \Sigma_l$ is the covariance matrix which is the inverse of the precision matrix $\cvec \Lambda_l$. Due to the observation model assumption $\cmat{H}=[\cmat{I},\cmat{0}]$, they take block diagonal form, $$\cvec \Sigma_l = \left[\begin{array}{cc}\cvec \Sigma_l^u & \cvec \Sigma^{s}_l \\ \cvec \Sigma^{s}_l & \cvec \Sigma^{l}_l  \end{array} \right], \textrm{ with } \cvec \Sigma_u = \textrm{diag}(\cvec \sigma_l^u), \; \cvec \Sigma_l = \textrm{diag}(\cvec \sigma_l^l) \textrm{ and } \cvec \Sigma_{s} = \textrm{diag}(\cvec \sigma_{l}^s).$$  

\paragraph{Proof:}

\paragraph{Case 1 (Single Observation):} Before deriving the update rule for $N$ conditionally iid observations, let us start with a simpler case consisting of a single observation $\cvec r$. If the marginal Gaussian distribution for the latent variable $\cvec l$ takes the form $ p(\mathbf{l}) =\mathcal{N}\left(\mathbf{l} \mid \boldsymbol{\mu}, \boldsymbol{\Lambda}^{-1}\right)$ and the conditional Gaussian distribution for he single observation $\cvec{r}$ given $\cvec{l}$ has the form
, $p(\mathbf{r} \mid \mathbf{l}) =\mathcal{N}\left(\mathbf{r} \mid \mathbf{H} \mathbf{l}+\mathbf{b}, \mathbf{L}^{-1}\right)$. Then the posterior distribution over $\mathbf{l}$ can be obtained in closed form as, 
\begin{equation}
\label{eq:bishop1}
\begin{aligned}
p(\mathbf{l} \mid \mathbf{r}) & =\mathcal{N}\left(\mathbf{l} \mid \mathbf{\Sigma}\left\{\mathbf{H}^{\mathrm{T}} \mathbf{L}(\mathbf{r}-\mathbf{b})+\boldsymbol{\Lambda} \boldsymbol{\mu}\right\}, \boldsymbol{\Lambda}^{-1}\right) ,\textrm{where } \boldsymbol{\Lambda}=\left(\boldsymbol{\Lambda}+\mathbf{H}^{\mathrm{T}} \mathbf{L} \mathbf{H}\right).
\end{aligned}
\end{equation}
We refer to Section 2.3.3 of \cite{bishop2006pattern}, to the proof for this standard result.

\paragraph{Case 2 (Set Of Observations):} Now instead of a single observation, we wish to derive a closed form solution for the posterior over latent variable $\boldsymbol{l}\in \mathbb{R}^{2d}$, given a set of N conditionally i.i.d observations $\Bar{r} = \{r_i\}_{i=1}^N$. Here each element $\boldsymbol{r_i} \in \mathbb{R}^d$ of the set $\Bar{r}$ is assumed to to have an observation model $\cmat{H}=[\cmat{I},\cmat{0}]$. In the derivation, we represent the set of N observations as a random vector $$\Bar{r} = \left[ \begin{array}{ll}
\cvec{r_1}  \\
\cvec{r_2} \\
\cvec{.} \\
\cvec{.} \\
\cvec{r_N}
\end{array}\right]_{Nd \times 1}.$$ 
Since each observation in the set $\Bar{r}$ are conditionally independent, we denote the conditional distribution over the context set as $\Bar{r}  \mid \mathbf{l} \sim \mathcal{N}\left(\bar{H} \mathrm{l},  \cmat{\Sigma}_{r}\right)$, where the diagonal covariance matrix has the following form: $$\cmat{\Sigma}_{r}= \left[ \begin{array}{llllll}
 \textrm{diag}(\cvec \sigma_{r_1}),&0,&0,&..,&0  \\
0,&\textrm{diag}(\cvec \sigma_{r_2}),&0,&..,&0 \\
\boldsymbol{.},&.,&.,&..,&. \\
\boldsymbol{. },&.,&.,&..,&.\\
0,&0,&0,&..,&\textrm{diag}(\cvec \sigma_{r_N})
\end{array}\right]_{Nd \times Nd}.
$$
The corresponding observation model  $\bar{\cmat H}$ is
$$
\bar{\cmat H} =  \left[ \begin{array}{ll}
\cmat{H}  \\
\cmat{H} \\
\cmat{.} \\
\cmat{.} \\
\cmat{H}
\end{array}\right]_{Nd \times 2d} = \left[ \begin{array}{ll}
\boldsymbol{I }, \boldsymbol{0 }  \\
\boldsymbol{I }, \boldsymbol{0} \\
\boldsymbol{.}, \boldsymbol{.} \\
\boldsymbol{. }, \boldsymbol{.} \\
\boldsymbol{I }, \boldsymbol{0}
\end{array}\right]_{Nd \times 2d}.$$

Now given the prior over the latent task variable $\mathrm{l} \sim \mathcal{N}\left(\mu_l^-, \boldsymbol{\Sigma}_l^-\right)$, the parameters of the posterior distribution over the task variable, $p(l|\Bar{r}) \sim \mathcal{N}\left(\mu_l^+, \boldsymbol{\Lambda}_l^+\right)$, can be obtained in closed-form substituting in Equation \eqref{eq:bishop1} as follows.
\begin{equation}
\begin{aligned}
\Lambda_l^+ & =  (\Sigma_l^+)^{-1}  \\ &= \boldsymbol{\Sigma}_l^{-1}+ \bar{\cmat {H}} ^T \cmat{\Sigma}_{r} \bar{\cmat {H}}  \\
& = \boldsymbol{\Sigma}_l^{-1}+ \left[ \begin{array}{ll}
 \textrm{diag}(\cvec \sigma_{r_1}),  \textrm{diag}(\cvec \sigma_{r_2}),  \textrm{diag}(\cvec \sigma_{r_3}), . , . ,   \textrm{diag}(\cvec \sigma_{r_N})  \\
\boldsymbol{0 }, \quad \quad \boldsymbol{0}, \quad \quad \boldsymbol{0}, \quad. , . ,\quad  \boldsymbol{0 }  \\
\end{array}\right]_{2d\times nd} \bar{ \cmat{H}} \\
& =\boldsymbol{\lambda}_l^{-} + \left[ \begin{array}{ll}
\mathrm{diag}(\sum_{i=1}^n\frac{1}{\cvec \sigma_{r_i}}), &\boldsymbol{0 }  \\
\boldsymbol{0 },& \boldsymbol{0} \\
\end{array}\right]_{2d\times 2d}  \\[15pt]
\mu_l^+ & = \boldsymbol{\mu}_l^{-} + (\Lambda^+)^{-1}\bar{\cmat {H}} ^T \left( \sigma_{\boldsymbol{r}}^{-2} \boldsymbol{I}\right)\left(\boldsymbol{y}-\bar{\cmat {H}}  \boldsymbol{\mu}_{\boldsymbol{x}}\right) \\
& = \boldsymbol{\mu}_l^{-}+ \Sigma^+ \bar{\cmat {H}} \left( \sigma_{\boldsymbol{r}}^{-2} \boldsymbol{I}\right) \left(\boldsymbol{y}- \bar{\cmat {H}}  \boldsymbol{\mu}_{\boldsymbol{x}}\right) \\
& = \boldsymbol{\mu}_l^{-}+ \Sigma^+\left[ \begin{array}{ll}
\sigma_{\boldsymbol{r_1}}^{-2}\boldsymbol{I }, \sigma_{\boldsymbol{r_2}}^{-2}\boldsymbol{I}, \sigma_{\boldsymbol{r_3}}^{-2}\boldsymbol{I }, . , . ,  \sigma_{\boldsymbol{r_n}}^{-2}\boldsymbol{I }  \\
\boldsymbol{0 }, \quad \quad \boldsymbol{0}, \quad \quad \boldsymbol{0}, \quad. , . ,\quad  \boldsymbol{0 }  \\
\end{array}\right] \left(\boldsymbol{y}-\bar{\cmat {H}}  \boldsymbol{\mu}_{\boldsymbol{x}}\right) \\
& =\boldsymbol{\mu}_l^{-} + \left[ \begin{array}{ll}
\boldsymbol{\sigma_l^{u+}}, &\boldsymbol{\sigma_l^{s+}}  \\
\boldsymbol{\sigma_l^{s+}} ,& \boldsymbol{\sigma_l^{l+}}  \\
\end{array}\right]  \left[ \begin{array}{l}
\sum_{n=1}^{N}  \left(\mathbf{r_n}-\mathbf{\mu}^{\mathrm{u},-}_l\right) \odot \frac{1}{\sigma_i} \\
\boldsymbol{0 } \\
\end{array}\right]  \\
& =\boldsymbol{\mu}_l^{-} + \left[ \begin{array}{l}
\boldsymbol{\sigma_l^{u+}} \\
\boldsymbol{\sigma_l^{s+}}  \\
\end{array}\right]  \odot \left[ \begin{array}{l}
\sum_{i=1}^{N}  \left(\mathbf{r_i}-\mathbf{\mu}^{\mathrm{u},-}_l\right) \odot \frac{1}{\cvec \sigma_{r_i}} \\
\sum_{i=1}^{N}  \left(\mathbf{r_n}-\mathbf{\mu}^{\mathrm{u},-}_l\right) \odot \frac{1}{\cvec \sigma_{r_i}} \\
\end{array}\right]  \\
\end{aligned}
\end{equation}

Here $\mu_l^+$ is the posterior mean and $\boldsymbol{\Lambda}_l^+$ is the posterior precision matrix.

\begin{coro}
\label{cor:1}
The closed form updates for the resulting posterior distribution $p(l|\Bar{r})$ is permutation invariant with respect to the observation set $\Bar{r}$.
\end{coro}

\subsection{Derivation For Matrix Inversions as Scalar Operations}

\begin{minv}
Consider a block  matrix of the following form $\cvec A = \left[\begin{array}{cc} \mathrm{diag}(\cvec a^u) & \mathrm{diag}(\cvec a^s) \\  \mathrm{diag}(\cvec a^s) &  \mathrm{diag}(\cvec a^l)  \end{array}\right]$. Then inverse $A^{-1} = \cvec B $ can be calculated using scalar operations and is given as, $\cvec B = \left[\begin{array}{cc} \mathrm{diag}(\cvec b^u) & \mathrm{diag}(\cvec b^s) \\  \mathrm{diag}(\cvec b^s) &  \mathrm{diag}(\cvec b^l)  \end{array}\right]$ where,
\begin{align}
\label{eq: matscalar}
\begin{split}
\cvec b^u & =\cvec a_l \oslash{ ( \cvec a_u \odot \cvec a_l- \cvec a_s \odot \cvec a_s )}  \\
\cvec b^s & = - \cvec a_s \oslash{ ( \cvec a_u \odot \cvec a_l- \cvec a_s \odot \cvec a_s )}  \\
\cvec b^l & =\cvec a_u \oslash{ ( \cvec a_u \odot \cvec a_l- \cvec a_s \odot \cvec a_s )} 
\end{split}
\end{align}.
\end{minv}

\paragraph{Proof:} To prove this we will use the following matrix identity of a partitioned matrix from \cite{bishop2006pattern}, which states
\begin{align}
\label{eq: matide}
\left(\begin{array}{ll}
\mathrm{A} & \mathrm{B} \\
\mathrm{C} & \mathrm{D}
\end{array}\right)^{-1}=\left(\begin{array}{cc}
\mathrm{M} & -\mathrm{MBD}^{-1} \\
-\mathrm{D}^{-1} \mathrm{CM} & \mathrm{D}^{-1}+\mathrm{D}^{-1} \mathrm{CMBD}^{-1}
\end{array}\right)
\end{align}
where M is defined as
$$
\mathrm{M}=\left(\mathrm{A}-\mathrm{BD}^{-1} \mathrm{C}\right)^{-1}.
$$ Here M is called the Schur complement of the Matrix on the left side of Equation \ref{eq: matide}. The algebraic manipulations to arrive at scalar operations in Equation \ref{eq: matscalar} are straightforward.

\newpage

\section{Additional Experiments and Plots}

\subsection{Additional results on ablation with discretization step $ H.\Delta t $}

\begin{wrapfigure}[12]{r}{.38\linewidth}
 \scalebox{0.75}{
\begin{subfigure}[b]{.38\textwidth}
    \centering
    \includegraphics[scale=0.5]{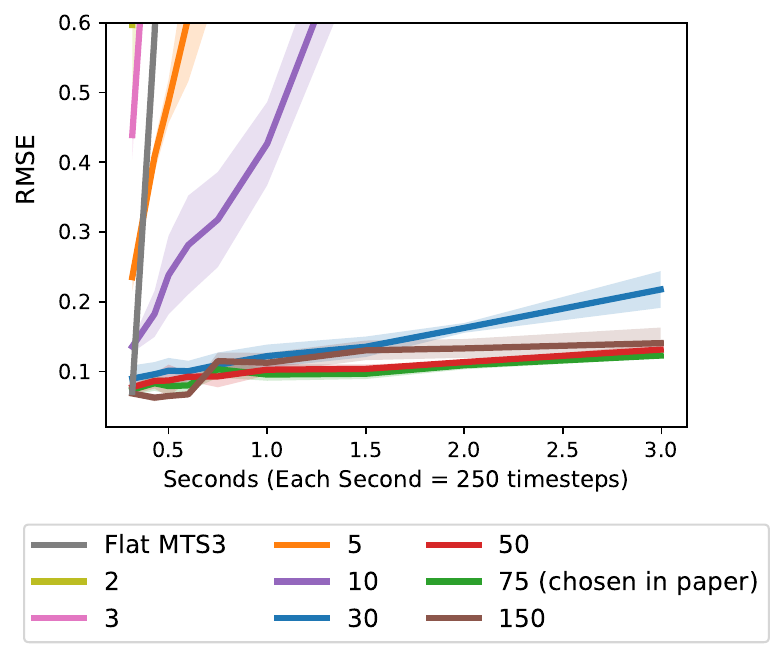}
     \end{subfigure}}
\caption{Ablation on discretization step $H. \Delta t$. The long-term prediction results in terms of RMSE, with different $H$ on the mobile dataset. } 
 \label{fig:mobAbl}
\end{wrapfigure}
In addition to the Hydraulics Dataset discussed in Section 6.4, we report the results of the ablation study with different values of $H.\Delta t$, for the mobile robot dataset. The higher the value of H, the slower the timescale of the task dynamics relative to the state dynamics. As seen in Figure \ref{fig:mobAbl}, smaller values of H (like 2,3,5 and 10) give significantly worse performance. Very large values of H (like 150) also result in degradation of performance. In the paper, we used a value of H=75.
\vfill
\subsection{Visualization of predictions given by different models.}

 In this section, we plot the multistep ahead predictions (mean and variance) by different models on 3 datasets on normalized test trajectories. Not that we omit NaN values in predictions while plotting.
 \newpage

 \subsubsection{Franka Kitchen}
  \begin{figure*}[ht!]
\begin{center}
\includegraphics[scale=0.75]{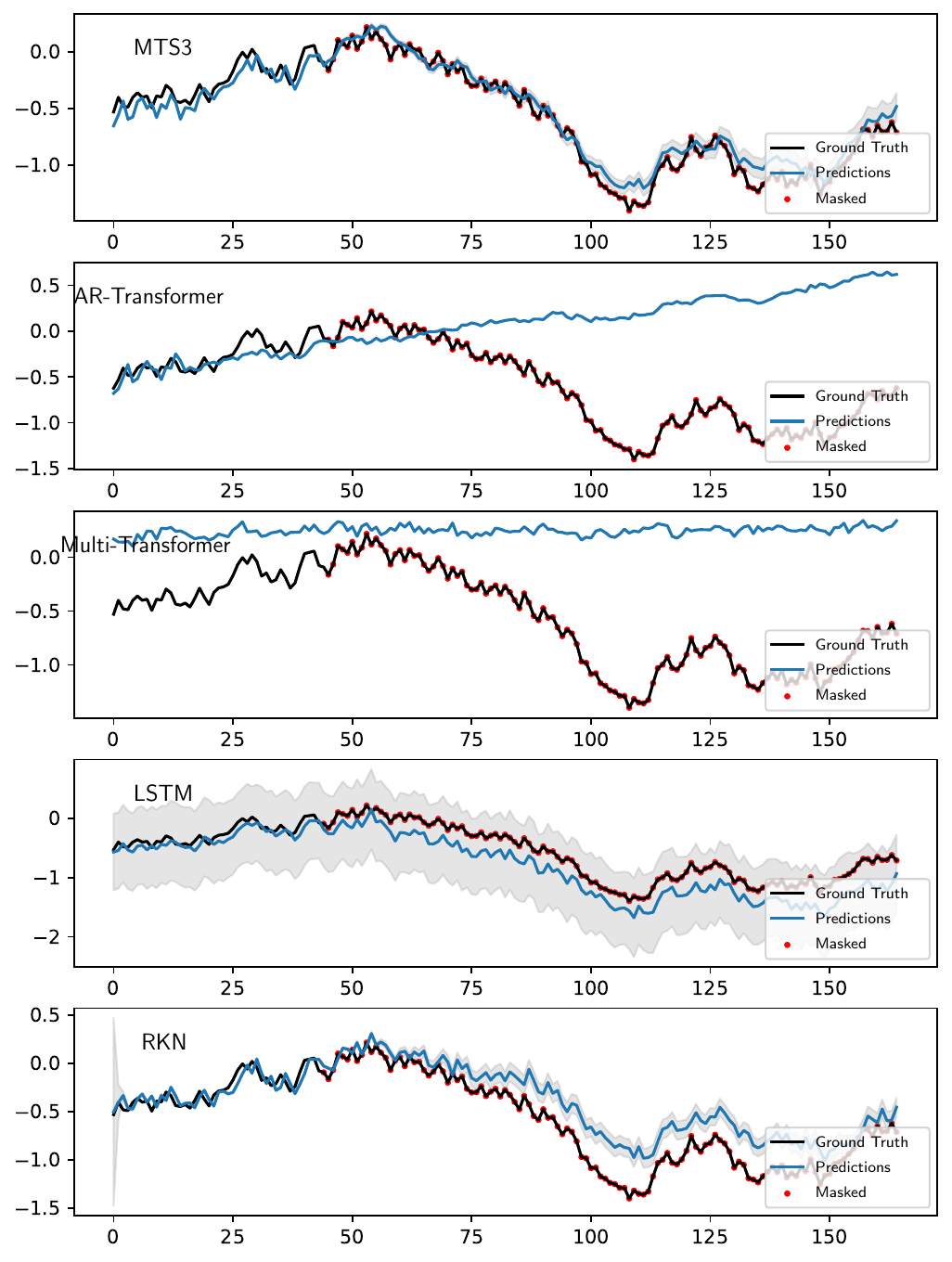}
\end{center}
\caption{Multi-step ahead mean and variance predictions for a particular joint (joint 1) of Franka Kitchen Environment. The multi-step ahead prediction starts from the first red dot, which indicates masked observations. MTS3 gives the most reliable mean and variance estimates.} 
 \label{fig:kitchen}
\end{figure*}
\newpage
 \subsubsection{Hydraulic Excavator}
 
\begin{figure*}[ht!]
\begin{center}
\includegraphics[scale=0.8]{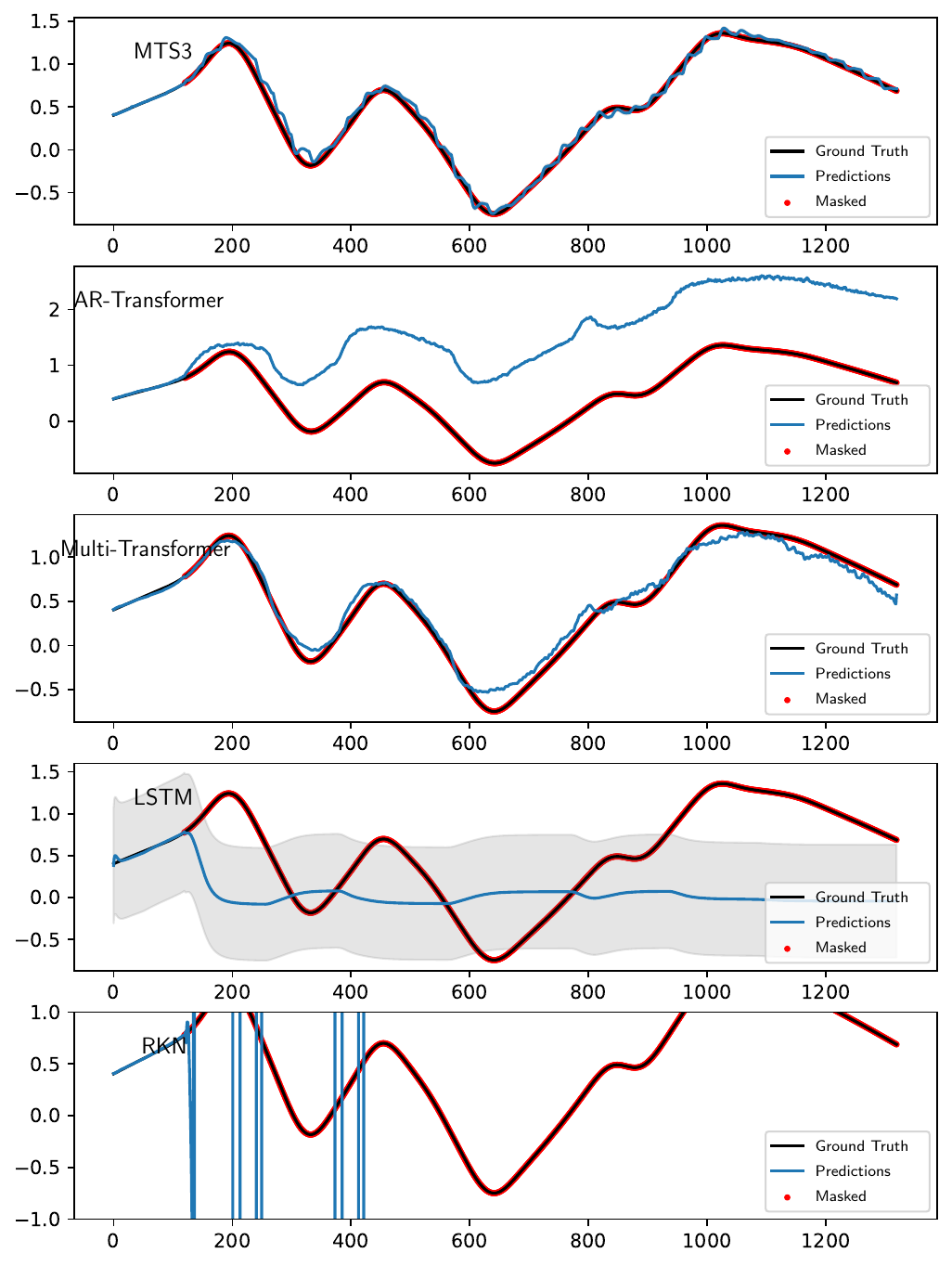}
\end{center}
\caption{Multi-step ahead mean and variance predictions for a particular joint (joint 1) of Excavator Dataset. The multi-step ahead prediction starts from the first red dot, which indicates masked observations. MTS3 gives the most reliable mean and variance estimates even up to 12 seconds into the future. Another interesting observation can also be seen in the predictions for MTS3, where after every window k of sts-SSM, which is 0.3 seconds (30 timesteps) long, the updation of the higher-level abstractions helps in grounding the lower-level predictions thus helping in the long horizon yet fine-grained predictions.} 
 \label{fig:exc}
\end{figure*}
\newpage

\subsubsection{Mobile Robot}
 
\begin{figure*}[ht!]
\begin{center}
\includegraphics[scale=0.8]{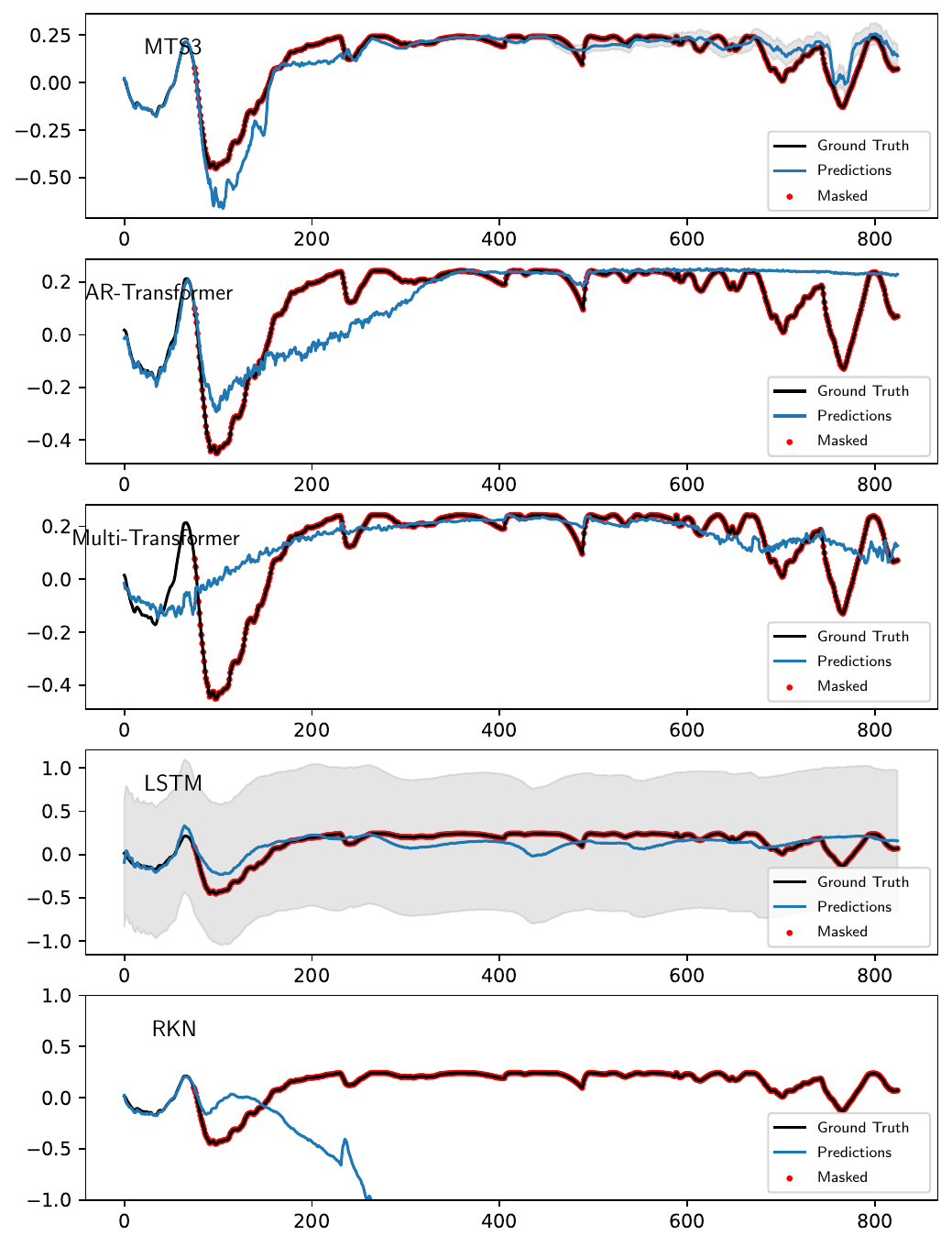}
\end{center}
\caption{Multi-step ahead mean and variance predictions for a particular joint (joint 7) of Mobile Robot Dataset. The multi-step ahead prediction starts from the first red dot, which indicates masked observations. MTS3 gives the most accurate mean and variance estimates among all algorithms. } 
 \label{fig:sin}
\end{figure*}
\newpage
\section{Robots and Data}
In all datasets, we only use information about agent/object positions and we mask out velocities to create a partially observable setting. All datasets are subjected to a mean zero, unit variance normalization during training. During testing, they are denormalized after predictions. The details of the different datasets used are explained below:

\subsection{D4RL Datasets}

\begin{figure*}[h]
\begin{center}
\includegraphics[scale=0.5]{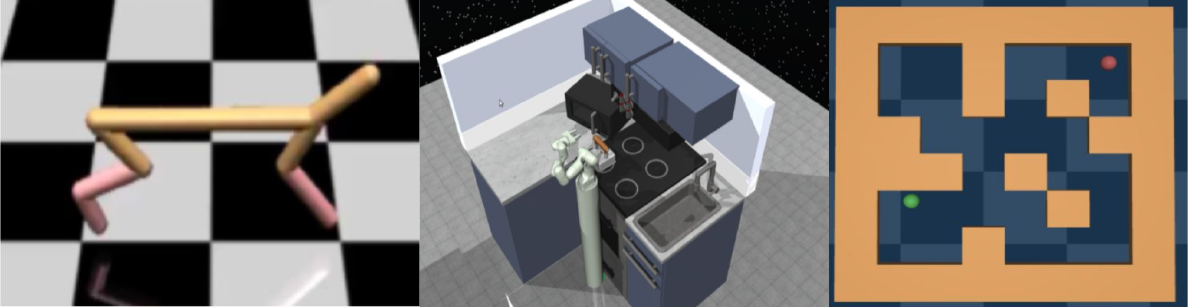}
\end{center}
\caption{ D4RL Environments: (left) HalfCheetah (middle) Franka Kitchen (right) Maze2D-Medium} 
 \label{fig:d4rl}
\end{figure*}

\textbf{Details:} We use a set of 3 different environments/agents from D4RL dataset~\cite{fu2020d4rl}, which includes the HalfCheetah, Franka Kitchen and Maze2D (medium) environment. \textbf{(a) HalfCheetah:} We used 1000 suboptimal trajectories collected from a policy trained to approximately 1/3 the performance of the expert. The observation space consists of 8 joint positions and the action space consists of 6 joint torques collected at 50 Hz frequency. 800 trajectories were used for training and 200 for testing. For the long horizon task, we used 1.2 seconds (60 timesteps) as context and tasked the model to predict 6 seconds (300 timesteps) into the future. \textbf{(b) Franka Kitchen:} The goal of the Franka Kitchen environment is to interact with the various objects to reach a desired state configuration. The objects you can interact with include the position of the kettle, flipping the light switch, opening and closing the microwave and cabinet doors, or sliding the other cabinet door. We used the "complete" version of the dataset and collected 1000 trajectories where all four tasks are performed in order. The observation space consists of 30 dimensions (9 joint positions of the robot and 21 object positions). The action space consists of 9 joint velocities clipped between -1 and 1 rad/s. The data was collected at a 50 Hz frequency. 800 trajectories were used for training and 200 for testing. For the long horizon task, we used 0.6 seconds (30 timesteps) as context and tasked the model to predict 2.7 seconds (135 timesteps) into the future. The dataset is complex due to multi-task, multi-object interactions in a single trajectory.
\textbf{(c) Medium Maze:} We used 20000 trajectories from a 2D Maze environment, where each trajectory consists of a force-actuated ball (along the X and Y axis) moving to a fixed target location. The observation consists of as the (x, y) locations and a 2D action space. The data is collected at 100 Hz frequency. 16000 trajectories were used for training and 4000 for testing. For the long horizon task, we used 0.6 seconds (60 timesteps) as context and tasked the model to predict 3.9 seconds (390 timesteps) into the future. Rendering of the three environments is shown in Figure \ref{fig:d4rl}.

\subsection{Hydraulic Excavator}

\begin{figure*}[h]
 \centering
\begin{subfigure}[b]{0.28\textwidth}
\includegraphics[scale=0.45]{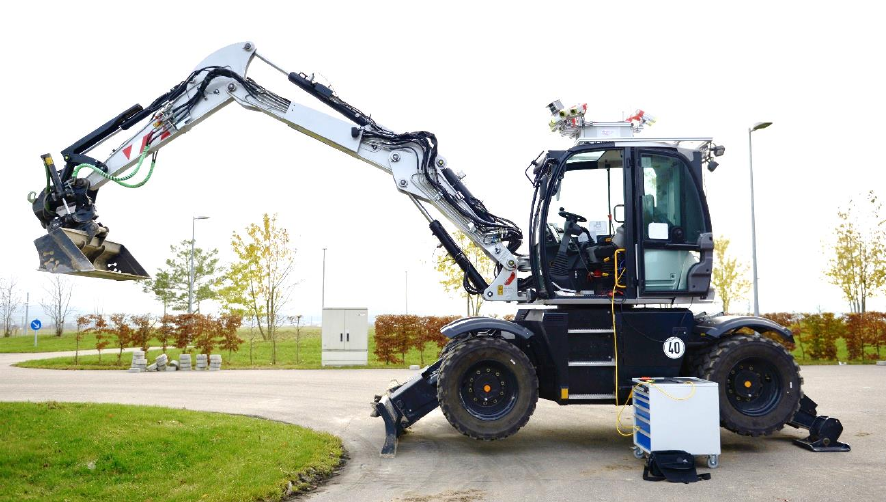}
\end{subfigure}
\hspace{3cm}
\begin{subfigure}[b]{0.28\textwidth}
\includegraphics[scale=0.2]{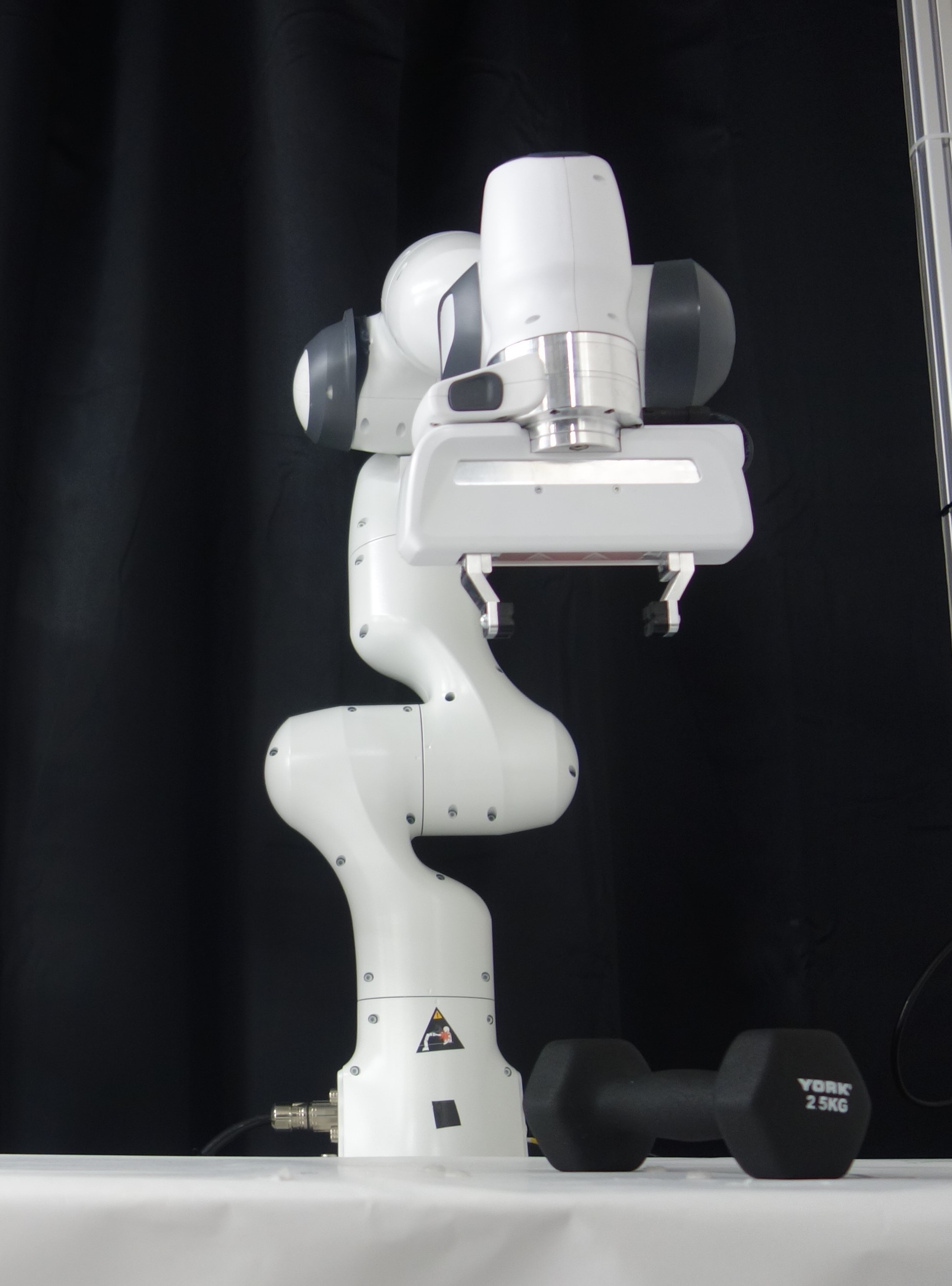}
\end{subfigure}
\caption{(left) JCB Hydradig 110W Excavator (right) Franka Emika Panda Robot} 
 \label{fig:exca}
\end{figure*}

\textbf{Details:} We collected the data from a wheeled excavator JCB Hydradig 110W show in Figure \ref{fig:exca}. The data was collected by actuating the boom and arm of the excavator using Multisine and Amplitude-Modulated Pseudo-Random Binary Sequence
(APRBS) joystick signals with safety mechanisms in place. A total of 150 mins of data was collected at a frequency of 100 Hz. of which was used as a training dataset and the rest as testing. The observation space consists of the boom and arm positions, while the joystick
signals are chosen as actions. For the long horizon task we used 1.5 seconds (150 timesteps) as context and tasked the model to predict 12 seconds (1200 timesteps) into the future.  \\  

\subsection{Panda Robot With Varying Payloads}

\textbf{Details:} We collected the data from a 7 DoF Franka Emika Panda manipulator during free motion and while manipulating loads with weights 0kg~(free motion), 0.5 kg, 1 kg, 1.5 kg, 2 kg and 2.5 kg. The robot used is shown in Figure \ref{fig:exca}. Data is sampled at a frequency of 100 Hz. The training trajectories were motions with loads of 0kg(free motion), 1kg, 1.5kg, and 2.5 kgs, while the testing trajectories contained motions with loads of 0.5 kg and 2 kg. The observations for the forward model consist of the seven joint angles in radians, and the corresponding actions were joint Torques in Nm. For the long horizon task we used 0.6 seconds (60 timesteps) as context and tasked the model to predict 1.8 seconds (180 timesteps) into the future.  \\  


\subsection{Wheeled Mobile Robot}
\label{app:wheel}
\begin{wrapfigure}[14]{r}{.5\linewidth}
\begin{subfigure}[b]{0.5\textwidth}
         \includegraphics[scale=0.2]{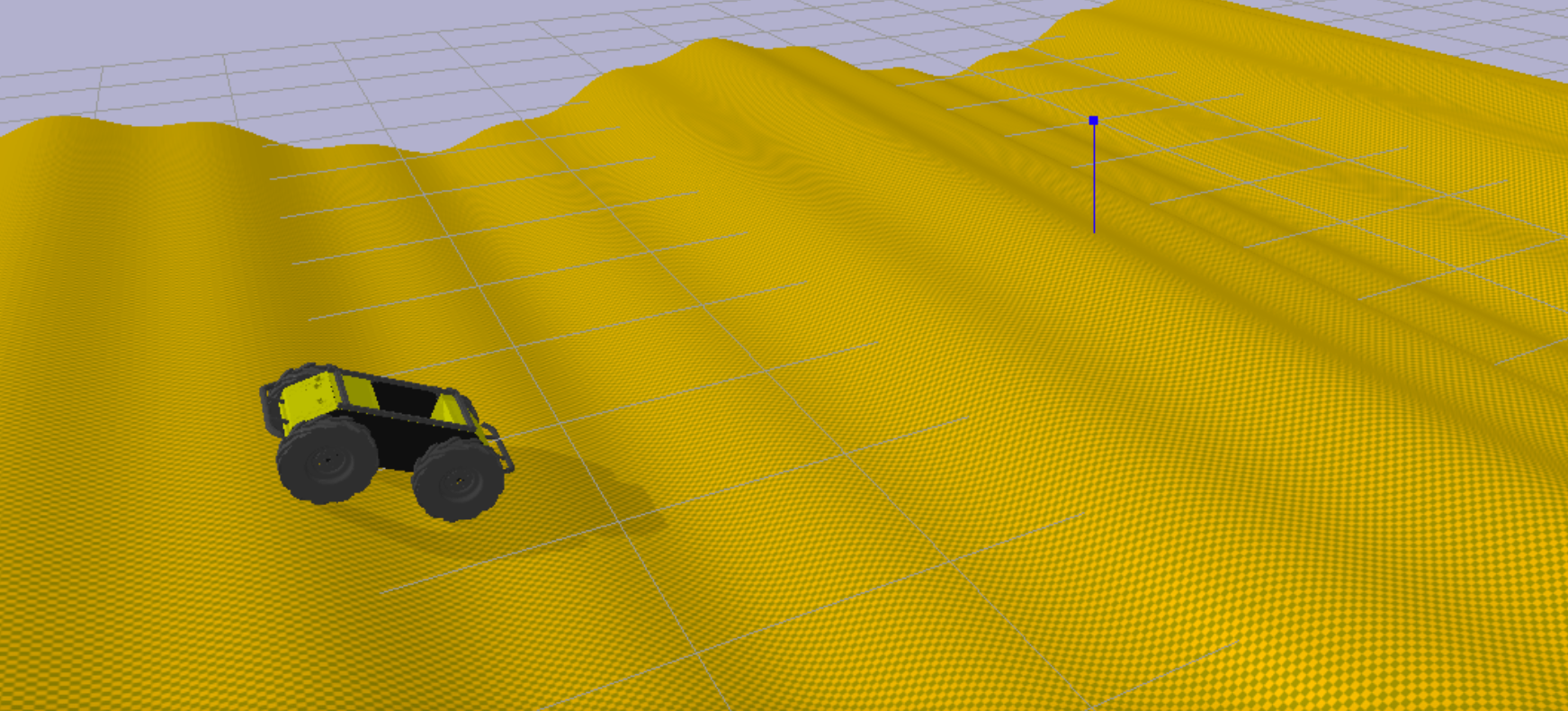}
     \end{subfigure}
\caption{Wheeled Mobile Robot traversing terrain with complex variations in slopes induced by a mix of sine functions.}
\label{fig:wheeled}
 \end{wrapfigure}
\textbf{Observation and Data Set:} We collected 50 random trajectories from a Pybullet simulator a wheeled mobile robot traversing terrain with slopes generated by a mix of sine waves as shown in Figure \ref{fig:wheeled}. Data is sampled at high frequencies~(500Hz). 40 out of the 50 trajectories were used for training and the rest 10 for testing. The observations consist of parameters which completely describe the location and orientation of the robot. The observation of the robot at any time instance $t$ consists of the following features:
$$
\begin{array}{r}
o_{t}=[x, y, z, \cos (\alpha), \sin (\alpha), \cos (\beta) \\
\sin (\beta), \cos (\gamma), \sin (\gamma)]
\end{array}
$$
where,
$x, y, z$ - denote the global position of the Center of Mass of the robot,
$\alpha, \beta, \gamma-$ Roll, pitch and yaw angles of the robot respectively, in the global frame of reference~\citep{sonker2020adding}.  For the long horizon task we used 0.6 seconds (150 timesteps) as context and tasked the model to predict 3 seconds (750 timesteps) into the future.   \\


\newpage

\section{Hyperparameters and Compute Resources}
\paragraph{Compute Resources}For training MTS3, LSTM, GRU and Transformer models we used compute nodes with (i) Nvidia 3090 and (ii) Nvidia 2080 RTX GPUs.
For training more computationally expensive locally linear models like RKN, HiP-RSSM we used
compute nodes with NVIDIA A100-40 GPUs.

\paragraph{Hyperparameters} Hyperparameters were selected via grid search. In general, the performance of MTS3 is not very sensitive to hyperparameters. Among all the baselines, Transformer models were most sensitive to hyperparameters (see Appendix E.5 for details of Transformer architecture). 

\paragraph{Discretization Step:} For MTS3, the discretization step for the slow time scale SSM as discussed in Section 3.1 for all datasets was fixed as $\cmat H \cdot \Delta t = 0.3$ seconds. In our experiments, we found that discretization values between $0.2 \leq\cmat H \cdot \Delta t \leq 0.5$ seconds give similar performance.

\paragraph{Rule Of thumb for choosing discretization step in MTS3:} For any N-level MTS3 as defined in Section 3.4, we recommend searching for discretization factor $H_i$ as a hyperparameter. However, as a general rule of thumb, it can be chosen as $H_i=(\sqrt[N]{T})^i$, where $T$ is the maximum prediction horizon required / episode length. This ensures that very long recurrences are divided between smaller equal-length task-reconfigurable local SSM windows (of length $\sqrt[N]{T}$) spread across several hierarchies.

\paragraph{Encoder Decoder Architecture:} For all recurrent models (MTS3, HiP-RSSM, RKN, LSTM and GRU) we use a similar encoder-decoder architecture across datasets. Small variations from these encoder-decoder architecture hyperparameters can still lead to similar prediction performance as reported in the paper.\\

\underline{Observation Set Encoder} (MTS3): 1 fully connected + linear output:
\begin{itemize}
	\item Fully Connected 1: 240, ReLU
\end{itemize} 

\underline{Action Set Encoder} (MTS3): 1 fully connected + linear output:
\begin{itemize}
	\item Fully Connected 1: 240, ReLU
\end{itemize} 

\underline{Observation Encoder} (MTS3, HiP-RSSM, RKN, LSTM, GRU): 1 fully connected + linear output:
\begin{itemize}
	\item Fully Connected 1: 120, ReLU
\end{itemize} 

\underline{Observation Decoder} (MTS3, HiP-RSSM, RKN, LSTM, GRU): 1 fully connected + linear output:
\begin{itemize}
	\item Fully Connected 1: 120, ReLU
\end{itemize} 

\underline{Control Model (Primitive Action Encoder)} (MTS3, HiP-RSSM, RKN):
1 fully connected + linear output:
\begin{itemize}
	\item Fully Connected 1: 120, ReLU
\end{itemize} 

The rest of the hyperparameters are described below:

\subsection{D4RL Datasets}

\subsubsection{Half Cheetah}
\underline{\textbf{Recurrent Models}}
\begin{table}[htbp]
  \centering
  \begin{tabular}{|c|c|c|c|c|c|}
    \hline
    \textbf{Hyperparameters} & \textbf{MTS3} & \textbf{HiP-RSSM} & \textbf{RKN} & \textbf{LSTM} & \textbf{GRU} \\
    \hline
    Learning Rate & 3e-3 & 1e-3 & 1e-3 &1e-3 &1e-3 \\ \hline
    Latent Observation Dimension &15 & 15 & 15 & 15 & 15  \\ \hline
Observation Set Latent Dimension (sts-SSM) & 15 & - & - & - & - \\ \hline
Latent State Dimension &30 & 30 & 30 & 45 & 45  \\ \hline
Latent Task Dimension &30 & 30 & - & - & -  \\ \hline
Latent Abstract Action Dimension (sts-SSM) & 30 & - & - & - & - \\ 
    \hline
  \end{tabular}
\end{table}
 
\underline{Transition Model} (HiP-RSSM, RKN):
number of basis: 32 

\begin{itemize}
    \item $\alpha(\cvec{z}_t)$: No hidden layers - softmax output
\end{itemize}

\underline{\textbf{Autoregressive Transformer Baseline}}\\
Learning Rate: 1e-5\\
Optimizer Used: Adam Optimizer \\
Embedding size: 96 \\
Number of Decoder Layers: 4 \\
Number Of Attention Heads: 4

\underline{\textbf{Multistep Transformer Baseline}}\\
Learning Rate: 1e-5\\
Optimizer Used: Adam Optimizer \\
Embedding size: 128 \\
Number Of Encoder Layers: 2\\
Number of Decoder Layers: 1 \\
Number Of Attention Heads: 4

\subsubsection{Franka Kitchen}
\underline{\textbf{Recurrent Models}}
\begin{table}[htbp]
  \centering
  \begin{tabular}{|c|c|c|c|c|c|}
    \hline
    \textbf{Hyperparameters} & \textbf{MTS3} & \textbf{HiP-RSSM} & \textbf{RKN} & \textbf{LSTM} & \textbf{GRU} \\
    \hline
    Learning Rate & 3e-3 & 9e-4 & 9e-4 &1e-3 &1e-3 \\ \hline
    Latent Observation Dimension &30 & 30 & 30 & 30 & 30  \\ \hline
Observation Set Latent Dimension (sts-SSM) & 30 & - & - & - & - \\ \hline
Latent State Dimension &60 & 60 & 60 & 90 & 90  \\ \hline
Latent Task Dimension &60 & 60 & - & - & -  \\ \hline
Latent Abstract Action Dimension (sts-SSM) & 60 & - & - & - & - \\
    \hline
  \end{tabular}
\end{table}
 
\underline{Transition Model} (HiP-RSSM, RKN):
number of basis: 15 

\begin{itemize}
    \item $\alpha(\cvec{z}_t)$: No hidden layers - softmax output
\end{itemize}

\underline{\textbf{Autoregressive Transformer Baseline}}\\
Learning Rate: 5e-5\\
Optimizer Used: Adam Optimizer \\
Embedding size: 64 \\
Number of Decoder Layers: 4 \\
Number Of Attention Heads: 4


\underline{\textbf{Multistep Transformer Baseline}}\\
Learning Rate: 1e-5\\
Optimizer Used: Adam Optimizer \\
Embedding size: 64 \\
Number Of Encoder Layers: 2\\
Number of Decoder Layers: 1 \\
Number Of Attention Heads: 4


\subsubsection{Maze 2D}
\underline{\textbf{Recurrent Models}}
\begin{table}[htbp]
  \centering
  \begin{tabular}{|c|c|c|c|c|c|}
    \hline
    \textbf{Hyperparameters} & \textbf{MTS3} & \textbf{HiP-RSSM} & \textbf{RKN} & \textbf{LSTM} & \textbf{GRU} \\
    \hline
    Learning Rate & 3e-3 & 9e-4 & 9e-4 &1e-3 &1e-3 \\ \hline
    Latent Observation Dimension &30 & 30 & 30 & 30 & 30  \\ \hline
Observation Set Latent Dimension (sts-SSM) & 30 & - & - & - & - \\ \hline
Latent State Dimension &60 & 60 & 60 & 90 & 90  \\ \hline
Latent Task Dimension &60 & 60 & - & - & -  \\ \hline
Latent Abstract Action Dimension (sts-SSM) & 60 & - & - & - & - \\
    \hline
  \end{tabular}
\end{table}
 
\underline{Transition Model} (HiP-RSSM, RKN):
number of basis: 15 

\begin{itemize}
    \item $\alpha(\cvec{z}_t)$: No hidden layers - softmax output
\end{itemize}

\underline{\textbf{Autoregressive Transformer Baseline}}\\
Learning Rate: 5e-5\\
Optimizer Used: Adam Optimizer \\
Embedding size: 96 \\
Number of Decoder Layers: 4 \\
Number Of Attention Heads: 4


\underline{\textbf{Multistep Transformer Baseline}}\\
Learning Rate: 1e-5\\
Optimizer Used: Adam Optimizer \\
Embedding size: 128 \\
Number Of Encoder Layers: 2\\
Number of Decoder Layers: 1 \\
Number Of Attention Heads: 4


\subsection{Franka Robot Arm With Varying Loads}
\underline{\textbf{Recurrent Models}}
\begin{table}[htbp]
  \centering
  \begin{tabular}{|c|c|c|c|c|c|}
    \hline
    \textbf{Hyperparameters} & \textbf{MTS3} & \textbf{HiP-RSSM} & \textbf{RKN} & \textbf{LSTM} & \textbf{GRU} \\
    \hline
    Learning Rate & 3e-3 & 9e-4 & 9e-4 &3e-3 &3e-3 \\ \hline
    Latent Observation Dimension &15 & 15 & 15 & 15 & 15  \\ \hline
Observation Set Latent Dimension (sts-SSM) & 15 & - & - & - & - \\ \hline
Latent State Dimension &30 & 30 & 30 & 45 & 45  \\ \hline
Latent Task Dimension &30 & 30 & - & - & -  \\ \hline
Latent Abstract Action Dimension (sts-SSM) & 30 & - & - & - & - \\ 
    \hline
  \end{tabular}
\end{table}

\underline{Transition Model} (HiP-RSSM,RKN):
number of basis: 32
\begin{itemize}
    \item $\alpha(\cvec{z}_t)$: No hidden layers - softmax output
\end{itemize}

\underline{\textbf{Autoregressive Transformer Baseline}}\\
Learning Rate: 5e-5\\
Optimizer Used: Adam Optimizer \\
Embedding size: 64 \\
Number of Decoder Layers: 4 \\
Number Of Attention Heads: 4


\underline{\textbf{Multistep Transformer Baseline}}\\
Learning Rate: 2e-5\\
Optimizer Used: Adam Optimizer \\
Embedding size: 64 \\
Number Of Encoder Layers: 2\\
Number of Decoder Layers: 1 \\
Number Of Attention Heads: 4


\subsection{Hydraulic Excavator}
\begin{table}[htbp]
  \centering
  \begin{tabular}{|c|c|c|c|c|c|}
    \hline
    \textbf{Hyperparameters} & \textbf{MTS3} & \textbf{HiP-RSSM} & \textbf{RKN} & \textbf{LSTM} & \textbf{GRU} \\
    \hline
    Learning Rate & 3e-3 & 8e-4 & 8e-4 &1e-3 &1e-3 \\ \hline
    Latent Observation Dimension &15 & 15 & 15 & 15 & 15  \\ \hline
Observation Set Latent Dimension (sts-SSM) & 15 & - & - & - & - \\ \hline
Latent State Dimension &30 & 30 & 30 & 45 & 45  \\ \hline
Latent Task Dimension &30 & 30 & - & - & -  \\ \hline
Latent Abstract Action Dimension (sts-SSM) & 30 & - & - & - & - \\ 
    \hline
  \end{tabular}
\end{table}

\underline{Transition Model} (HiP-RSSM,RKN):
number of basis: 15
\begin{itemize}
    \item coefficient net $\alpha(\cvec{z}_t)$: No hidden layers - softmax output
\end{itemize}

\underline{\textbf{Autoregressive Transformer Baseline}}\\
Learning Rate: 1e-5\\
Optimizer Used: Adam Optimizer \\
Embedding size: 96 \\
Number of Decoder Layers: 4 \\
Number Of Attention Heads: 4


\underline{\textbf{Multistep Transformer Baseline}}\\
Learning Rate: 5e-5\\
Optimizer Used: Adam Optimizer \\
Embedding size: 64 \\
Number Of Encoder Layers: 2\\
Number of Decoder Layers: 1 \\
Number Of Attention Heads: 4


\subsection{Wheeled Robot Traversing Uneven Terrain}
\begin{table}[htbp]
  \centering
  \begin{tabular}{|c|c|c|c|c|c|}
    \hline
    \textbf{Hyperparameters} & \textbf{MTS3} & \textbf{HiP-RSSM} & \textbf{RKN} & \textbf{LSTM} & \textbf{GRU} \\
    \hline
    Learning Rate & 3e-3 & 8e-4 & 8e-4 &1e-3 &1e-3 \\ \hline
    Latent Observation Dimension &30 & 30 & 30 & 30 & 30  \\ \hline
Observation Set Latent Dimension (sts-SSM) & 30 & - & - & - & - \\ \hline
Latent State Dimension &60 & 60 & 60 & 90 & 90  \\ \hline
Latent Task Dimension &60 & 60 & - & - & -  \\ \hline
Latent Abstract Action Dimension (sts-SSM) & 60 & - & - & - & - \\ 
    \hline
  \end{tabular}
\end{table}
 
\underline{Transition Model} (HiP-RSSM,RKN):
number of basis: 15
\begin{itemize}
    \item coefficient net $\alpha(\cvec{z}_t)$: No hidden layers - softmax output
\end{itemize}

\underline{\textbf{Autoregressive Transformer Baseline}}\\
Learning Rate: 5e-5\\
Optimizer Used: Adam Optimizer \\
Embedding size: 128 \\
Number of Decoder Layers: 4 \\
Number Of Attention Heads: 4


\underline{\textbf{Multistep Transformer Baseline}}\\
Learning Rate: 5e-5\\
Optimizer Used: Adam Optimizer \\
Embedding size: 64 \\
Number Of Encoder Layers: 4\\
Number of Decoder Layers: 2 \\
Number Of Attention Heads: 4


\subsection{Transformer Architecture Details}
For the AR-Transformer Baseline, we use a GPT-like autoregressive version of transformers except that for the autoregressive input we also concatenate the actions to make action conditional predictions.

For Multi-Transformer we use the same direct multistep prediction and loss as in recent Transformer time-series forecasting literature~\cite{zhou2021informer,liu2022nonstationary,nie2023a,zeng2022transformers}. A description of the action conditional direct multi-step version of the transformer is given in Algorithm \ref{algo:trans}.

\begin{algorithm}
\caption{MultiStep Transformer}
\textbf{Require:} Input past observations $\mathbf{o_{inp}} \in \mathbb{R}^{S \times C}$; Input Past Actions $\mathbf{a_{inp}} \in \mathbb{R}^{S \times A}$;  Future Actions $\mathbf{a_{pred}} \in \mathbb{R}^{O \times A}$;Input Length $S$; Predict length $O$; Observation Dimension $C$; Action Dimension $A$; Feature dimension $d_k$; Encoder layers number $N$; Decoder layers number $M$. \\
\vspace{0.5cm}
1: $ \mathbf{o_{inp}} \in \mathbb{R}^{S \times C}, \cmat a_{inp} \in \mathbb{R}^{S \times A} , \cmat a_{pred} \in \mathbb{R}^{O \times A} $   \\
2: $\cmat X_{inp} = \mathrm{ConCatFeatureWise
}\left(\cmat o_{inp}, \cmat a_{inp}\right)$ \hspace{3.4cm} $\triangleright \cmat X_{inp} \in \mathbb{R}^{S \times (C+A)}$ \\
3: $\cmat X_{pred} = \mathrm{ConCatFeatureWise
}\left(\operatorname{Zeros}(O, C), \cmat a_{pred}\right)$ \hspace{2cm} $\triangleright \cmat X_{pred} \in \mathbb{R}^{O \times (C+A)}$\\
4: $\mathbf{X}_{\mathrm{enc}}, \mathbf{X}_{\mathrm{dec}}=\mathbf{X}_{\mathrm{inp}}, \operatorname{ConCat}\left(\cmat X_{inp}, \cmat X_{pred}\right)  \quad \quad \triangleright \mathbf{X}_{\mathrm{enc}} \in \mathbb{R}^{S \times (C+A)}, \mathbf{X}_{\mathrm{dec}} \in \mathbb{R}^{\left(S+O\right) \times (C+A)}$ \\
5: $\cmat X_{\mathrm{enc}}^{0}=\operatorname{Embed}\left(\mathrm{X}_{\mathrm{enc}}\right)$ 
\hspace{7cm} $\triangleright \mathbf{X}_{\mathrm{enc}}^{0} \in \mathbb{R}^{S \times d_k}$ \\
6: \For{$l$ in $\{1, \cdots, N\}$}{
    7: $\quad \cmat X_{\mathrm{enc}}^{l-1}=\operatorname{LayerNorm}\left(\cmat X_{\mathrm{enc}}^{l-1}+\operatorname{Attn}\left(\cmat X_{\mathrm{enc}}^{l-1}\right)\right)$ 
    \hspace{3.1cm} $\triangleright \cmat X_{\mathrm{enc}}^{l-1} \in \mathbb{R}^{S \times d_k}$ \\
    8: $\quad \cmat X_{\text {enc }}^{l }=\operatorname{LayerNorm}\left(\cmat X_{\text {enc }}^{l-1 }+\operatorname{FFN}\left(\cmat X_{\text {enc }}^{l-1 }\right)\right)$ 
    \hspace{3cm} $ \triangleright \cmat X_{\mathrm{enc}}^{l } \in \mathbb{R}^{S \times d_k}$
}
9: $\cmat X_{\mathrm{dec}}^{0}=\operatorname{Embed}\left(\cmat X_{\mathrm{dec}}\right)$
\hspace{3cm} $\triangleright \cmat X_{\mathrm{dec}}^{0} \in \mathbb{R}^{\left(S+O\right) \times d_k}$\\
10: \For{\textbf{for} $l$ in $\{1, \cdots, M\}$}{
    11: $\quad \cmat X_{\operatorname{dec}}^{l-1}=\operatorname{LayerNorm}\left(\cmat X_{\operatorname{dec}}^{l-1}+\operatorname{Attn}\left(\cmat X_{\operatorname{dec}}^{l-1 }\right)\right)$ 
    \hspace{2cm} $\triangleright$ Decoder \\
    12: $\quad \cmat X_{\mathrm{dec}}^{l-1 }=\operatorname{LayerNorm}\left(\cmat X_{\mathrm{dec}}^{l-1 }+\operatorname{Attn}\left(\cmat X_{\mathrm{dec}}^{l-1 \prime}, \cmat X_{\mathrm{enc}}^{N }\right)\right)$ 
    $\triangleright \cmat X_{\mathrm{dec}}^{l-1 } \in \mathbb{R}^{\left(S+O\right) \times d_k}$ \\
    13:$\quad \cmat X_{\mathrm{dec}}^{l }=\operatorname{LayerNorm}\left(\cmat X_{\mathrm{dec}}^{l-1 }+\operatorname{FFN}\left(\cmat X_{\mathrm{dec}}^{l-1 }\right)\right)$ 
    \hspace{1cm} $\triangleright \cmat X_{\mathrm{dec}}^{l } \in \mathbb{R}^{\left(S+O\right) \times d_k}$
}
14: $\mathbf{y}=\operatorname{MLP}\left(\cmat X_{\mathrm{dec}}^{M}\right)$
\hspace{0.5cm} $\triangleright \mathbf{y} \in \mathbb{R}^{(S+O) \times C}$ \\
15: Return y \hspace{7cm}$\triangleright$ Return the prediction results
\label{algo:trans}
\end{algorithm}

\newpage

\section{Limitations}
We list some of the limitations of the paper here. (i) We restricted our definition and experiments to MTS3 with two levels of temporal abstractions, which was sufficient in many of our tasks. However, for certain tasks like the Maze2D, we believe more hierarchies can help. As discussed in the main paper the method and inference scheme allows easy addition of more Feudal~\cite{dayan1992feudal} hierarchies with larger discretization steps ($\cmat H \cdot \Delta t$). (ii) We restrict our application to action conditional long horizon future predictions and do not use the model for (hierarchical) control. A probabilistically principled formalism for hierarchical control as an inference problem, that builds upon MTS3 models is left for future work.  (iii) Finally, we restrict our experiments to proprioceptive sensors from the agent and objects. The performance of MTS3 which relies on ``reconstruction loss'' as the objective is yet to be validated on noisy high dimensional sensor inputs like Images. Image-based experiments and ``non-reconstruction'' based losses~\cite{lecun2022path} can be taken up as future work.

\section{Broader Impact}
While we do not foresee any immediate negative societal impacts of our work, we do believe that machines that can replicate human intelligence at some point should be able to reason at multiple levels of temporal abstractions using internal world models~\cite{lecun2022path}. Having intelligent agents with type 2 reasoning capabilities can have both positive and negative impacts. We believe identifying and mitigating the potentially harmful effects of such autonomous systems is the responsibility of sovereign governments.

\end{document}